\documentclass[10pt,journal,compsoc]{IEEEtran}
\usepackage{amsmath,amsfonts}
\usepackage{algorithm}

\usepackage{array}

\usepackage[caption=false,font=normalsize,labelfont=sf,textfont=sf]{subfig}

\usepackage{textcomp}
\usepackage{stfloats}
\usepackage{url}
\usepackage{verbatim}
\usepackage{graphicx}
\usepackage{cite}
\usepackage[table,xcdraw]{xcolor}
\usepackage{bbding}
\usepackage{shadowtext}
\usepackage{colortbl} 
\usepackage{booktabs}
\usepackage{makecell}
\usepackage{bm}
\usepackage{caption}
\usepackage{subcaption}
\usepackage{amssymb} 
\usepackage[colorlinks,linkcolor=red,anchorcolor=blue,citecolor=green]{hyperref}
\usepackage{pifont}
\usepackage{multirow} 
\usepackage{bbding}
\usepackage{orcidlink}

\usepackage{marvosym}
\usepackage[most]{tcolorbox}

\newcommand{\toolns}{\emph{LanEvil++}}
\newcommand{\tool}{\toolns\space}
\newcommand{\advlmns}{ADVLMs}
\newcommand{\advlm}{\advlmns\space}

\newcommand{\revised}[1]{\textcolor{black}{#1}}

\newcommand{\ie}{\textit{i}.\textit{e}.}
\newcommand{\eg}{\textit{e}.\textit{g}.} 
\newcommand{\Tref}[1]{Tab.~\ref{#1}}

\newcommand{\Fref}[1]{Fig.~\ref{#1}}
\newcommand{\Sref}[1]{Sec.~\ref{#1}}

\begin{document}

\title{Benchmarking the Robustness of Autonomous Driving to Environmental Illusions: \\A Lane Perception Perspective}

\author{Tianyuan Zhang$^{\orcidlink{0000-0001-9874-6828}}$, Xianglong Liu\Letter$^{\orcidlink{0000-0001-8425-4195}}$, Aishan Liu$^{\orcidlink{0000-0002-4224-1318}}$, Lu Wang$^{\orcidlink{0009-0004-1733-4178}}$, Yitong Zhang$^{\orcidlink{0009-0000-1138-4503}}$, Peng Yue$^{\orcidlink{0009-0004-7679-5388}}$, \\Mingchuan Zhang, Siyuan Liang$^{\orcidlink{0000-0002-6154-0233}}$, Dacheng Tao$^{\orcidlink{0000-0001-7225-5449}}$,~\IEEEmembership{Fellow,~IEEE}

\thanks{T. Zhang,  X. Liu (\Letter corresponding author), A. Liu, L. Wang and Y. Zhang are with SKLCCSE, the School of Computer Science and Engineering, Beihang University, Beijing 100191, China.}
\thanks{P. Yue is with the School of Cyber Science and Technology, Sun Yat-sen University, Shenzhen 518107, China.}
\thanks{M. Zhang is with Henan University of Science and Technology, Luoyang 471023, China.}
\thanks{S. Liang is with the School of Computing, National University of Singapore, Singapore 117417.}
\thanks{D. Tao is with the College of Computing \& Data Science, Nanyang Technological University, Singapore 639798.}
}
        

\markboth{Journal of \LaTeX\ Class Files,~Vol.~14, No.~8, August~2021}%
{Shell \MakeLowercase{\textit{et al.}}: A Sample Article Using IEEEtran.cls for IEEE Journals}

\IEEEtitleabstractindextext{%
\begin{abstract}
Environmental illusions (\eg, shadows, reflections, and tire marks) are naturally existing yet overlooked phenomena in real-world driving environments. They can disturb visual perception, leading to misinterpretation of the scene and posing serious safety risks to autonomous driving (AD) systems. However, existing researches largely overlook these phenomena, leaving a critical gap.
To address this issue, we study AD robustness through the lane perception perspective, a fundamental task supporting core functions like cruise control and lane centering. We focus on two representative models: conventional lane detection (LD) and vision-language model-based systems (\advlmns).
In this work, we introduce the first benchmark, \toolns, for evaluating the robustness of lane perception under environmental illusions. \tool encompasses 14 types of illusions and leverages the CARLA simulator to generate 94 high-fidelity, fully controllable 3D scenes, yielding a dataset of 90,292 annotated images, 1,596 video clips, and 41,855 visual question answering pairs.
Extensive evaluations demonstrate that environmental illusions substantially degrade the performance of state-of-the-art LD methods. On average, LD models experience a \revised{5.27\%} drop in Accuracy and a \revised{10.49\%} decline in F1-score, while \advlm show a \revised{2.03\%} reduction in GPT-score and a \revised{0.75\%} drop in Language-score. Among all illusions, shadows emerge as the most disruptive factor, reducing accuracy by up to \revised{7.20\%}. Furthermore, closed-loop simulations using OpenPilot and LMDrive reveal that these illusions can lead to incorrect driving decisions, underscoring their real-world implications. Complementary real-world case studies highlight safety-critical failures in actual traffic scenes. 
To enhance robustness, we propose the Multimodal Illusion Defense Approach (MIDA), which uses hard examples to improve illusion resistance. MIDA achieves substantial gains under challenging conditions, boosting robustness by 4.23\% on LD models and 3.82\% on \advlmns. We hope this work brings greater attention to the threats posed by environmental illusions and motivates the development of more robust AD systems. Part of our dataset and demos can be found at the {\href{https://tianyuan2001.github.io/lanevilpp.github.io/}{\textcolor{blue}{https://tianyuan2001.github.io/lanevilpp.github.io/}}}.

\end{abstract}

\begin{IEEEkeywords}
Autonomous driving, environmental illusion, robustness benchmark.
\end{IEEEkeywords}
}

\maketitle

\section{Introduction}
\label{sec:intro}

\begin{figure}[!t]
    \includegraphics[width=0.95\linewidth]{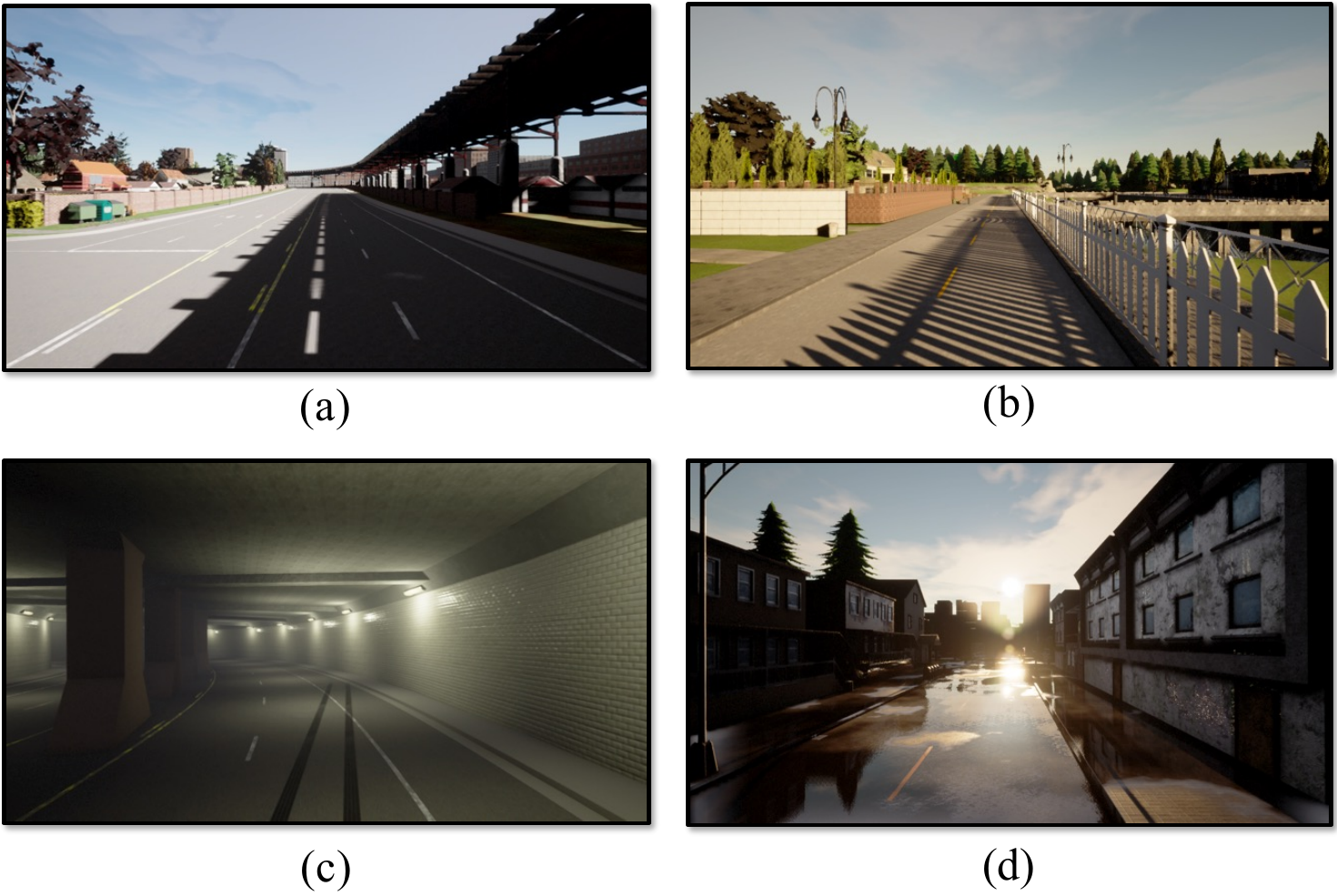}
    \caption{Illustration of naturally existing yet overlooked environmental illusions (\eg, shadows). Misinterpreting these visually deceptive but physically patterns can lead to incorrect perception, posing potential safety risks for AD systems.}
    \label{fig:headimage}
\end{figure}

\IEEEPARstart{E}{nvironmental} illusions are \emph{naturally existing yet overlooked} phenomena that can mislead autonomous driving (AD) systems. In real-world traffic scenarios, environmental illusions such as shadows and reflections, as shown in \Fref{fig:headimage}, frequently appear but remain insufficiently addressed. 
\revised{Unlike adverse weather, motion blur, or synthetic noise that primarily degrade global visual quality, environmental illusions introduce natural yet structurally misleading patterns that resemble lane boundaries or road markings.}
These natural corruptions pose serious risks to the safety of AD systems by increasing their vulnerability and potentially endangering human lives and property \cite{Uber_Crash}. Nevertheless, the impact of environmental illusions on AD models remains largely unexplored. Given the safety-critical nature of AD, thoroughly evaluating its robustness to such illusions is essential before real-world deployment.

Lane perception is a fundamental task in AD, responsible for identifying lane lines and road edges \cite{qin2020ultra, zheng2021resa, qin2022ultra}, and forms the basis for key functions such as lane centering and adaptive cruise control \cite{hillel2014recent}. It is typically approached through two representative methods: conventional lane detection (LD) and vision-language model-based AD systems (\advlmns). LD relies on pixel-level supervision to produce accurate geometric predictions, while ADVLM combines visual input with language guidance to enable reasoning and multi-modal understanding.
Although lane perception models have shown strong performance on standard traffic datasets (\eg, CULane \cite{pan2018spatial_SCNN}, DriveLM-nuScenes \cite{sima2024drivelm}), their robustness to environmental illusions remains largely unexplored.

To address this, we investigate the robustness of AD models to environmental illusions from the perspective of lane perception.
Unlike common corruption datasets \cite{hendrycks2019benchmarking_imagenet-c,dong2023benchmarking_3dob} (\eg, snow), which introduce uniform noise across the entire input, environmental illusions involve localized, case-specific alterations that appear natural, making them more difficult to identify and address. 
Although environmental illusions naturally occur in real-world scenarios, collecting large-scale data with diverse types of these illusions remains challenging. 
Simulation testing, known for its cost-efficiency and reproducibility, has become a widely used alternative to real-world experiments, particularly in AD research \cite{alghodhaifi2021simulation_1, rosique2019simulation_2, fremont2020simulation_3}.
Therefore, we employ a simulation testing pipeline \cite{li2022metadrive, Li_2023_CVPR, wang2021dual} to generate synthetic images \cite{richter2016playing_for_data, richter2017playing_for_benchmark, sekkat2020omniscape}. 
Particularly, we first design 3D environments in the CARLA simulator \cite{dosovitskiy2017carla}, then apply perturbations to target objects (\eg, roads) with environmental illusions, and finally render the corrupted images.

Based on the pipeline, we rigorously analyze the influence factors in lane perception including \textit{Dynamic Objects}, \textit{Static Facilities}, and \textit{Environmental Conditions}, and systematically design 14 types with 5 severity levels that collectively address a wide spectrum of environmental illusions on the road and propose the \tool robustness evaluation benchmark.
Overall, our \tool benchmark consists of two parts. For conventional LD models, \tool includes 94 editable 3D environments and a dataset of 90,292 sampled images, comprising 40,000 clean training images and 50,292 test images. For \advlmns, \tool defines 6 distinct task categories tailored to lane perception scenarios, including 1,596 clips and 41,855 QA pairs.

Leveraging \toolns, we conducted extensive experiments to benchmark the robustness of lane perception models. For conventional LD models, we observed significant performance degradation, with an average drop of \revised{5.27\%} in Accuracy and \revised{10.49\%} in F1-score. Among all illusion types, \texttt{shadow} had the most severe impact, reducing Accuracy by \revised{7.20\%}. For \advlmns, we similarly observed notable degradation, with GPT Score decreasing by \revised{2.03\%} and Language Score by \revised{0.75\%} on average. The \texttt{Traffic Obstruction} illusion had the greatest effect, causing a \revised{3.51\%} drop in GPT Score. To enhance model robustness against environmental illusions, we propose the Multimodal Illusion Defense Approach (MIDA), a novel noise defense baseline that leverages hard examples for lane perception models. MIDA achieves a 4.23\% improvement on conventional LD models and a 3.82\% improvement on \advlm compared to vanilla detectors under these illusions. To better demonstrate the potential of \toolns, we conducted closed-loop experiments on AD systems both in the simulation world (OpenPilot \cite{openPilot} and LMDrive \cite{shao2024lmdrive}), where we observed incorrect decisions resulting in vehicle accidents. Finally, we conducted case studies on real-world scenarios containing environmental illusions, which demonstrate the threats in practice. We hope this work will raise awareness regarding potential security threats \cite{liang2023badclip,liu2023pre,liu2023does,liang2024poisoned,liang2024vl,li2024semantic} in AD scenarios. 

A preliminary version of this work~\cite{zhang2024lanevil} was accepted as a poster at ACM MM 2024. \ding{182} \textbf{Expanded Research Scope}. While the earlier work focused on environmental illusions in conventional LD models, the current study targets the broader AD system. We adopt lane perception as a representative lens to systematically analyze illusion-induced vulnerabilities in AD. \ding{183} \textbf{Benchmark Augmentation.} The original \textit{lanEvil} benchmark has been significantly expanded and restructured. We added new annotations and tasks (\eg, Scene Description and Object Recognition), enabling the evaluation of \advlmns.
\ding{184} \textbf{Methodological Advancements.} We improve the AAM method by introducing AAM++, which enhances robustness by incorporating both high/low attention regions from hard examples. Additionally, we propose PAT++, a specialized prompt-tuning scheme adapted for \advlmns, improving their resilience to environmental illusions in multimodal settings.
\ding{185} \textbf{Comprehensive Experiments.} We evaluated not only conventional LD models on \toolns, but also several state-of-the-art \advlm (\eg, LMDrive, Dolphins), demonstrating that environmental illusions pose significant challenges to both types of models.

Our \textbf{contributions} are as follows:

\begin{itemize}

    \item To our knowledge, this is the first study to investigate the impact of environmental illusions on the robustness of AD from the lane perception perspective.

    \item We developed the \tool benchmark, comprising 14 typical environmental illusions, 90,292 images, 94 editable 3D cases, and 1,596 clips with 41,855 QA pairs, supporting user customization.

    \item Extensive experiments show that environmental illusions significantly degrade performance in both conventional LD models and \advlmns. \texttt{Shadow} caused the largest drop in LD Accuracy (\revised{-7.20\%}), while \texttt{Traffic Obstruction} led to the greatest impact on \advlmns, reducing GPT Score by \revised{3.51\%}.
    

    \item We propose MIDA to improve the robustness of lane perception models against environmental illusions. It includes AAM++ for conventional LD, which augments hard examples with mixed attention regions, and PAT++ for \advlmns, a prompt-tuning scheme enhancing robustness in multimodal settings.
    
    \item Compared to our previous work~\cite{zhang2024lanevil}, this study expands the scope from conventional LD to the entire AD, from the perspective of lane perception to highlight the threat of environmental illusions.
  
\end{itemize}
\section{Related Work}

\subsection{Lane Detection} \label{sec:lane_detection_models}


LD refers to the process of identifying lane markings or road boundaries.
In recent years, deep learning-based approaches have become the dominant framework for LD, leveraging their capacity to autonomously learn intricate feature representations and spatial patterns.
In general, LD methods can be systematically categorized into five major paradigms:
\ding{182} \emph{segmentation-based methods} \cite{neven2018towards, pan2018spatial_SCNN, liu2020heatmap}, which formulate LD as a pixel-level semantic segmentation problem;
\ding{183} \emph{keypoint-based methods} \cite{ko2021key_PINet, qu2021focus_FOLOLane, wang2022keypoint_GANet}, which detect sparse but discriminative keypoints along lane markings and subsequently cluster them into coherent lane structures through geometric or learned association rules;
\revised{\ding{184} \emph{anchor-based methods} \cite{li2019line_LineCNN, tabelini2021keep_LaneATT, su2021structure_SGNet, han2022laneformer, zheng2022clrnet}, where LD is framed as an anchor-driven regression task, leveraging predefined reference points to efficiently localize and parameterize lane boundaries;}
\ding{185} \emph{row-wise classification methods} \cite{yoo2020end_E2E-LMD, hou2020inter_IntRA-KD, qin2020ultra,qin2022ultra}, which adopt a vertical scanning strategy, sequentially predicting the most probable lane-containing regions for each image row to reconstruct complete lane markings; 
\ding{186} \emph{parameter-based methods} \cite{tabelini2021polylanenet, liu2021end_LSTR, chen2023bsnet, feng2022rethinking}, which reformulate LD as an optimization problem by directly estimating the parametric representations (\eg, polynomial curves).


In this study, we conduct a comprehensive benchmarking analysis to evaluate and compare the robustness of all aforementioned LD methodologies.

\subsection{VLMs for Autonomous Driving}

The integration of \advlm has attracted substantial research interest, primarily due to its multimodal processing prowess. This integration holds the promise of bolstering the performance of AD systems in real-world scenarios.

Early AD research explored the combination of LLMs with driving tasks. For instance, DriveLikeHuman \cite{fu2024drive} emulated human driving learning processes using LLMs, while GPT-Driver \cite{mao2023gptdriver} and Agent-Driver \cite{mao2023language} utilized GPT-3.5 for motion planning. Nevertheless, these studies predominantly relied on linguistic input, neglecting the integration of rich visual features.

Subsequently, an increasing number of studies have been dedicated to exploring \advlmns. Recent works, such as Reason2Drive \cite{nie2024reason2drive}, LingoQA \cite{marcu2023lingoqa}, and Dolphins \cite{ma2024dolphins}, have concentrated on the Visual Question Answering (VQA) capabilities of VLMs in AD. These studies have probed into how VLMs can enhance scene comprehension, behavior prediction, and dialogue in driving-related tasks, highlighting the models' reasoning and explanatory abilities crucial for informed driving decisions.

In the domain of driving planning and control, several studies have made notable contributions. GPT-Driver \cite{mao2023gptdriver}, Driving with LLMs \cite{chen2024driving}, and MTD-GPT \cite{liu2023mtdgpt} were among the first to apply VLMs to driving planning. However, their open-loop operation failed to account for cumulative errors and end-to-end interpretability issues. LMDrive \cite{shao2024lmdrive} represented a significant advancement by introducing a VLM-based driving method in closed-loop settings, enabling continuous feedback for more accurate decision-making.

\revised{Furthermore, methods like DriveLM \cite{sima2024drivelm}, DriveMLM \cite{wang2023drivemlm}, OmniDrive \cite{wang2025omnidrive}, and DriveGPT4 \cite{xu2024drivegpt4} have integrated VQA with driving planning and control within VLM frameworks.} These approaches transcend basic interactions, implementing more refined driving control and decision-making reasoning by leveraging VLMs to analyze complex driving scenarios and generate sophisticated driving strategies.

Despite these achievements, the safety of \advlm remain significant concerns. Deep neural networks, including VLMs, encounter difficulties when handling out-of-distribution data \cite{kong2023robob3d, kong2023robodepth, xie2025benchmarking}. In AD, the inability to manage rare or unexpected scenarios can pose severe safety risks.

\subsection{Lane Perception Benchmarks} 
The advancement of lane perception technologies has been largely driven by the availability of benchmark datasets. Existing benchmarks generally fall into two categories: conventional LD benchmarks and those designed for \advlmns.

\textbf{
Conventional LD Benchmarks
}
Traditional LD benchmarks have enabled significant progress by offering datasets under increasingly diverse and challenging conditions. 
Initial datasets are notably limited in scope, exemplified by the CalTech dataset \cite{aly2008real} with its collection of 1,224 frames captured in standard urban environments under consistent weather conditions.
More recent benchmarks demonstrate significantly greater sophistication. 
VPGNet \cite{lee2017vpgnet} substantially expands the complexity with 20,000 images capturing challenging urban traffic situations, 
while TuSimple \cite{tusimple} specializes in highway scenarios with precise lane annotations. 
Additionally, contemporary datasets incorporate even more varied driving conditions: 
CULane \cite{pan2018spatial_SCNN} provides an extensive set of over 130,000 images where approximately 72.3\% present difficult scenarios including congested traffic and intense glare; 
BDD-100K \cite{yu2020bdd100k} encompasses a wide spectrum of illumination conditions and six distinct severe weather patterns; 
LLAMAS \cite{behrendt2019unsupervised_llamas} exhibits variable lane marking density that correlates with observation distance and spatial location; 
furthermore, CurveLane \cite{xu2020curvelane} specifically targets the detection of curved lane geometries.

Despite their scale, the aforementioned standard datasets are not designed with robustness evaluation in mind.
Dedicated \textbf{robustness benchmarks} examining natural corruptions in LD scenarios remain scarce.
Existing pioneering robustness benchmarks primarily focus on evaluating the robustness of image classification \cite{hendrycks2019benchmarking_imagenet-c,zhang2021interpreting, guo2023towards, liu2020spatiotemporal, xiao2023robustmq, xiao2023benchmarking} and object detection \cite{hendrycks2019benchmarking_imagenet-c,liu2023exploring, liu2022harnessing, liu2021training, tang2021robustart, zhang2023benchmarking, jiang2023exploring, zhang2024towards, wangattack, zhangenhancing, liang2021generate, liang2020efficient, wei2018transferable,liang2022parallel,liang2022large,wang2023diversifying,liu2023x,he2023generating,liu2023improving,he2023sa,muxue2023adversarial,lou2024hide} against common perturbations including blur effects, adverse weather conditions, and digital artifacts.
Recent work has begun exploring perception robustness for autonomous driving systems under common corruptions \cite{dong2023benchmarking_3dob,kong2023robo3d,ying2026safebench,xiao2026detoxifying,ying2025jailbreak,liu2025pre,liu2025compromising}, though these investigations predominantly address general 3D perception tasks like detection and segmentation, with the simulated corruptions (e.g., motion blur) not being specifically tailored for LD.
Furthermore, while some studies have developed specialized lane-like adversarial attacks for autonomous driving systems \cite{sato2021dirty, boloor2020attacking_blackline}, such approaches fall outside the purview of our current investigation.

In summary, while current LD datasets have made significant progress in evaluating method performance, a systematic examination of scenario-specific corruptions, particularly shadows and reflections, remains conspicuously absent. Our \tool addresses this critical gap by introducing a comprehensive benchmark specifically designed for robustness evaluation against LD corruptions.

\textbf{\advlm Benchmarks}
With the growing interest in applying \advlm tasks, several benchmarks have been proposed \cite{sun2020scalability, ma2024dolphins, sima2024drivelm, shao2024lmdrive}.
For example, nuScenesQA \cite{qian2024nuscenesqa} provides a wide array of driving scenarios, accompanied by rich sensor data and annotations, which is conducive to evaluating the perception and prediction capabilities of VLMs. BDD100K \cite{yu2020bdd100k} focuses on heterogeneous multitask learning, offering a large-scale dataset for diverse driving-related tasks. The Waymo Dataset \cite{sun2020scalability}, renowned for its high-quality and extensive data, is advantageous for training and assessing advanced VLMs. Additionally, datasets like DriveLM \cite{sima2024drivelm} integrate language-based annotations with traditional driving datasets, facilitating language-driven perception and decision-making in AD.

Regarding evaluation metrics, traditional metrics such as ROUGE \cite{lin2004rouge}, BLEU \cite{papineni2002bleu}, and CIDEr \cite{vedantam2015cider}, originally developed for NLP tasks like text summarization and machine translation, are applied to assess language-driven interactions in AD. Modern GPT-based scoring methods \cite{chen2024driving, sima2024drivelm} harness the power of large language models to score model responses in AD tasks.

Critically, the robustness of \advlm in LD tasks remains an open problem. Existing benchmarks do not assess model performance under long-tail or corrupted lane scenarios, nor do they evaluate failure modes specific to language-guided LD.
Safety-critical edge cases, such as occluded or low-visibility lanes, are rarely emphasized.
To mitigate these issues, safety-oriented benchmarks like DriveBench \cite{xie2025are} have been proposed. DriveBench evaluates VLM reliability across multiple settings, including different corruptions and text-only inputs. It assesses 12 VLMs on four driving tasks using various metrics, aiming to expose VLM limitations and promote reliable AD, but it does not explicitly target LD.

This work aims to advance research on the robustness of LD in \advlmns, further promoting the systematic study of LD perception robustness in real-world AD scenarios.
\section{The \tool Dataset}

\subsection{Problem Definition}
\label{sec:environment_define} 

\quad\textbf{Lane detector.} 
A lane detection model, formalized as a parametric function \(f_{\bm{\theta}}(\mathbf{I}) \rightarrow \mathbf{loc} \in \mathbb{N}^K\) with learnable parameters \({\bm{\theta}}\), processes an input image  \(\mathbf{I} \in [0, 255]^3\) and generates \(K\) predicted lane positions \(\mathbf{loc}\). 
The learning objective is mathematically expressed as:

\begin{equation}  
\min_{\bm{\theta}} \mathbb{E}_{(\mathbf{I}, \mathbf{loc}_{gt}) \sim \mathbb{D}} \mathcal{L}(f_{\bm{\theta}}(\mathbf{I}), \mathbf{loc}_{gt}),  
\end{equation}  


\noindent where \(\mathbb{D}\) represents the training dataset, and \(\mathcal{L}(\cdot)\) denote the objective function that quantifies the discrepancy between the predicted lane localization and the corresponding ground-truth annotations \(\mathbf{loc}_{gt}\). 

\textbf{\advlmns.} An ADVLM \(F_{\bm{\theta}}(\mathbb{I}, t) \rightarrow \mathbf{res} \in \mathbb{R}\) with parameters \(\bm{\theta}\) takes a query \((\mathbb{I}, t)\) as input, where \(\mathbb{I} \subset \mathbb{I}_\text{all}\) is a sequence of visual inputs consisting of multiple frames rather than a single image, and \(t \in \mathbb{T}\) is a textual input. 
In the context of lane detection, the response $\mathbf{res}$ contains structured signals such as lane presence indicators, lane topology, or even continuous lane representations \(L \in \mathbf{res}\).
The training objective is formulated as:

\begin{equation}  
\min_{\bm{\theta}} \mathbb{E}_{((\mathbb{I}, t), \mathbf{res}_{gt}) \sim \mathbb{Q}} \mathcal{L}( F_{\bm{\theta}}(\mathbb{I}, t), \mathbf{res}_{gt}),  
\end{equation}  

\noindent where \(F_{\bm{\theta}}\) represents the probability function, \(\mathbf{res}_{gt}\) is the ground truth response, \(\mathbb{Q}\) is the input query domain consisting of visual inputs \(\mathbb{I} \subset \mathbb{I}_\text{all}\) and textual inputs \(t \in \mathbb{T}\), and \(\mathbb{R}\) is the response domain.

\textbf{Environment.} 
In practical deployment, the AD system initially captures the surrounding physical environment $\mathbf{\Phi}$ through its onboard sensors. 
Subsequently, the perceived 3D scene is transformed into a 2D image representation $\mathbf{I}=R(\mathbf{\Phi})$, which serves as the input for downstream processing. 
Here, $R(\cdot)$  denotes the environmental projection function that encodes the 3D-to-2D mapping.
Specifically, the environmental elements pertinent to AD scenarios can be broadly categorized into two classes: static infrastructure components $\mathbb{S}= \{{\mathbf{s}}_{1}, {\mathbf{s}}_{2}, ...,{\mathbf{s}}_{n}\}$ (\eg, road surfaces, traffic barriers) and dynamic entities $\mathbb{X}= \{{\mathbf{x}}_{1}, {\mathbf{x}}_{2}, ...,{\mathbf{x}}_{m}\}$ (\eg, moving pedestrians, surrounding vehicles).
Furthermore, environmental conditions $\mathbf{C}$ including illumination and meteorological factors, constitute additional influential variables. Consequently, the comprehensive environmental representation $\mathbf{\Phi}$ should be formally expressed as:

\begin{equation}
\label{equ:environment}
\Phi = (<\mathbb{S}, \mathbb{X}>,\mathbf{C}).
\end{equation}


The input image representation 
$\mathbf{I}$ for the AD  system should be synthesized from the environment $\mathbf{\Phi}$, incorporating defined configurations of both static and dynamic elements under specified conditions, formally expressed as:

\begin{equation}
\label{equ:environment_image}
\mathbf{I} = R(<\mathbb{S}, \mathbb{X}>,\mathbf{C}).
\end{equation}

\textbf{Environmental illusions on LD models.} 
The performance of AD systems may degrade due to inherent variations in the static infrastructure, dynamic entities, and environmental conditions, which introduce distortions in the synthesized input representation $\mathbf{I}$.
To systematically generate environmental illusions, we deliberately manipulate the constituent parameters of the environmental model $\Phi$ in Equation \ref{equ:environment}, thereby obtaining a modified environment representation $\hat{\Phi}$.
The resulting rendered image $\hat{\mathbf{I}}=R(\mathbf{\hat{\Phi}})$ exhibits localized perturbations that introduce carefully crafted environmental illusions in specific spatial regions.
Consequently, the performance of LD models, denoted by $f_{\bm{\theta}}(\hat{\mathbf{I}})$/$F_{\bm{\theta}}(\hat{\mathbb{I}},t)$, may experience measurable degradation. 
Our methodological focus in this work centers on systematically designing single-factor changes to generate corresponding environmental illusions.

\begin{figure*}[!t]
    \centering
    \includegraphics[width=0.93\linewidth]{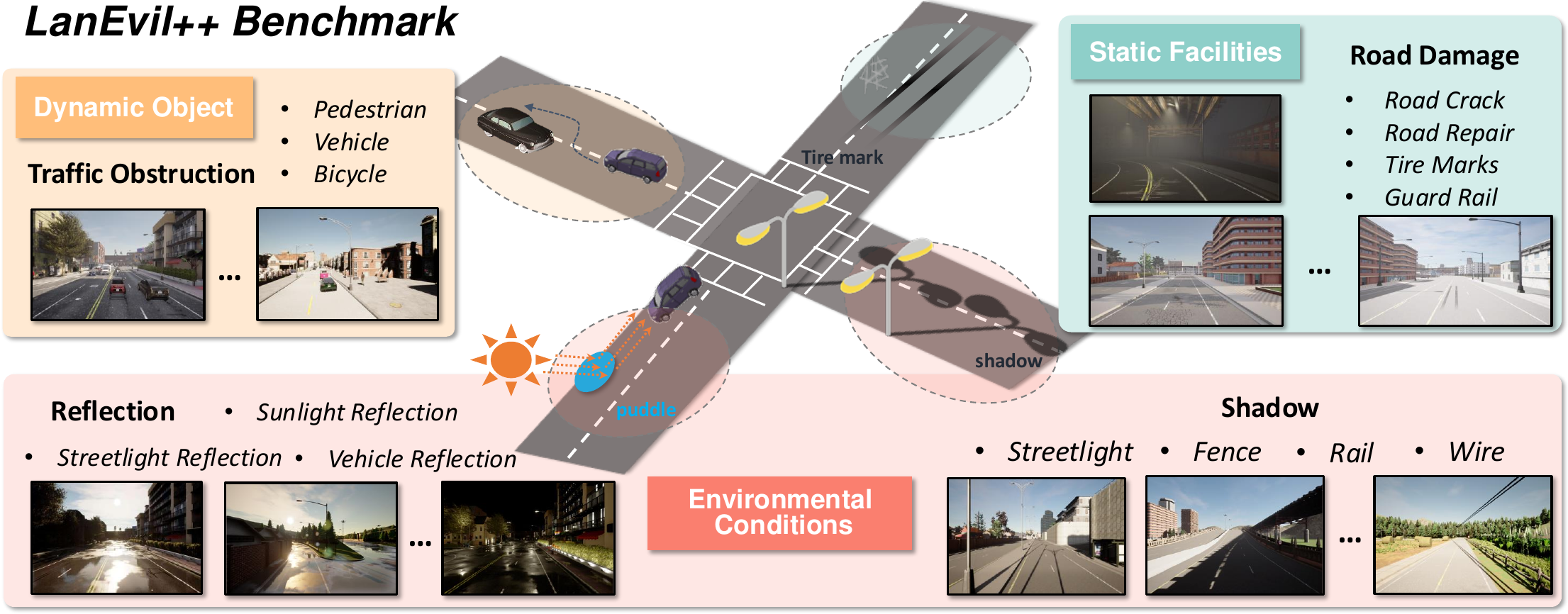}
    \caption{The framework of our \tool benchmark, which contains 14 specially-designed environmental illusion types from 4 categories, including road damage, traffic obstruction, reflection, and shadow.}
    \label{fig:framework}

\end{figure*}

\subsection{\revised{Illusion Category Selection}}
\label{sec:illusion_selection}

\revised{To ensure that the selected illusion categories are grounded in real-world safety relevance, we examined national crash databases that explicitly encode environmental contributory factors, including the U.S. FARS/CRSS system~\cite{NHTSA_2019_FARS_CRSS_Manual} and the U.K. DfT Road Safety Statistics~\cite{DfT_RoadSafety_Tables}. Both datasets report \textit{shadows}, \textit{dazzling reflections}, and \textit{visibility obstruction} as causal factors in traffic accidents, highlighting their tangible safety impact.
Complementing these cause-coded sources with the occurrence frequency observed in large-scale driving datasets (CULane~\cite{pan2018spatial_SCNN}, BDD100K~\cite{yu2020bdd100k}, LLAMAS~\cite{behrendt2019unsupervised_llamas}) and the failure ratios recorded in closed-loop lane-perception logs (OpenPilot~\cite{openPilot}), we derive a unified danger level to rank illusion categories by their combined prevalence and risk significance.
This analysis serves as the empirical foundation for the subsequent illusion design in \Sref{sec:corruption_types}, where these high-risk visual factors are systematically parameterized and synthesized under simulation conditions. \textit{A detailed comparison is provided in Supplementary Materials.}}

\subsection{Environmental Illusion Design}
\label{sec:corruption_types}

As illustrated in \Fref{fig:framework}, our proposed benchmark systematically incorporates four distinct categories of environmental illusions, comprising a total of 14 specific variants. Subsequently, we elaborate on the methodological design principles underlying each categorical implementation.
\revised{
Each illusion type is implemented with five controllable severity levels to emulate progressive real-world variations in visual intensity (\eg, illumination, occlusion, and structural complexity). 
\textit{Detailed parameter definitions for these severity levels are provided in the Supplementary Materials.}
}




\ding{182} 
The road damage effects correspond to the perturbations introduced to the static infrastructure components $\mathbb{S}$ within the roadway environment. Specifically, we formulate four representative categories of environmental illusions: \texttt{Road Crack}, \texttt{Road Repair}, \texttt{Tire Marks}, and \texttt{Guard Rail}. Notably, the \texttt{Road Crack} simulation incorporates a comprehensive spectrum of crack formations, spanning from localized transverse and longitudinal fractures to extensive interconnected fissure networks.
The repaired road segments present distinctive visual characteristics, with certain instances demonstrating linear features resembling lane markings; we formally designate this phenomenon as the \texttt{Road Repair} illusion. Additionally, we introduce the \texttt{Tire Marks} illusion to simulate vehicular skid patterns resulting from emergency braking maneuvers or collision scenarios.
Furthermore, we investigate potential misidentification scenarios where \texttt{Guard Rail} structures may be erroneously identified as lane markings, thereby compromising the accurate detection of road boundary delineations.


For a given static object $\mathbf{s}_{i}$ (\eg, road surface) within the environmental model $\mathbf{\Phi}$, we systematically alter its visual characteristics by physically applying the aforementioned perturbation patterns to targeted regions. 
This transformation yields the modified object representation $\hat{\mathbf{s}}_{i}=\mathbf{s}_{i}+g(\bm{\delta})$, where $g(\cdot)$ represents the perturbation transformation function and $\bm{\delta}$ encapsulates both the intensity parameters and spatial mask defining the perturbation.
Furthermore, we introduce new static objects $\mathbf{s}_{j}$ (\eg, guard rails) along road boundaries within the environmental model $\mathbf{\Phi}$. 
Consequently, the environmental illusions induced by Road Damage  can be mathematically represented through perturbing or adding new objects to the static infrastructure components, expressed as 
$\hat{\mathbb{S}} = \{\hat{{\mathbf{s}}}_{1}, ...,\hat{\mathbf{s}}_{i}, \mathbf{s}_{i+1}, \cdots, \mathbf{s}_{i+N}\}$
where $N$ denotes the number of newly incorporated objects.




\ding{183} \textbf{Traffic Obstruction.} 
This category of environmental illusions stems from the impact of dynamic entities $\mathbb{X}$ within roadway environments. 
As critical components in LD scenarios, traffic participants demand rigorous safety considerations and particularly robust detection mechanisms.
This study primarily examines three fundamental types of traffic participants: \texttt{Pedestrian}, \texttt{Vehicle}, and \texttt{Bicycle}. These dynamic elements may occlude critical lane markings and consequently degrade LD performance. 
Furthermore, we systematically vary participant densities to simulate diverse traffic scenarios with varying complexity levels.


For each experimental scenario, we generate $n$ distinct instances $\mathbf{x}_{i}$ corresponding to the specified illusion type. 
Unlike the Road Damage scenario where static infrastructure is modified, we initialize the dynamic object set $\mathbb{X}$ as  $\varnothing$ and subsequently introduce traffic participants at strategic spatial locations throughout the roadway. 
Formally, the Traffic Obstruction illusion can be represented through the augmentation of $m$ dynamic entities in the environment:
$\hat{\mathbb{X}} = \{{\mathbf{x}}_{1}, \mathbf{x}_{2}, \mathbf{x}_{3}, \cdots ,{\mathbf{x}}_{m}\}$.

\ding{184} \textbf{Shadow.} 
Environmental conditions $\mathbf{C}$, including meteorological and illumination variations, significantly influence the visual perception of roadway surfaces and associated infrastructure, thereby inducing additional environmental illusions. 
A particularly impactful factor involves temporal variations in illumination conditions, which generate dynamic shadow patterns across road surfaces. These shadows (\eg, from guardrails or streetlights) often exhibit morphological similarities to authentic lane markings, potentially compromising LD model performance.
To illustrate, during twilight conditions, the pronounced solar elevation angle produces extensive shadow coverage, whereas midday illumination with high intensity and minimal solar angle yields sharply defined shadow boundaries. 
Accounting for these photometric variations, we formulate four distinct categories of shadow-based environmental illusions: \texttt{Streetlight}, \texttt{Fence}, \texttt{Rail}, and \texttt{Wire}. 
In particular, the \texttt{Streetlight} and \texttt{Fence} illusions generate substantially elongated shadow patterns on pavement surfaces under low-angle solar illumination.
Furthermore, we introduce a novel shadow deception phenomenon induced by the semi-transparent structural components of \texttt{Rail}. 
Through careful modulation of illumination angles, these railings produce luminous linear projections on road surfaces that exhibit striking geometric and chromatic similarities to authentic lane markings, thereby generating particularly challenging visual ambiguities. 
Our investigation also reveals analogous effects with \texttt{Wire} infrastructure, where both the opaque shadow regions and the semi-luminous inter-wire spaces may be misinterpreted as legitimate lane demarcations.


Building upon this analytical framework, we synthesize shadow-induced illusions through systematic modulation of illumination parameters in the simulated environment. 
The resulting perturbed images $\mathbf{I}^{(l, a)}$ are derived from Equation \ref{equ:environment_image}, where the hyperparameters ${(l, a)}$ respectively correspond to luminance intensity and angular configuration of the light source.





\ding{185} \textbf{Reflection.} 
Meteorological variations, such as rainfall, would lead to the formation of reflective water accumulations on road surfaces. 
These specular reflections adversely affect lane marking detection accuracy. 
Furthermore, image acquisition under backlit conditions introduces significant photometric distortions due to heightened reflective interference.
Building upon these observations, we systematically formulate three types of illusion Reflection illusions: \texttt{Sunlight Reflection}, \texttt{Streetlight Reflection}, and \texttt{Vehicle Reflection}. 
The \texttt{Sunlight Reflection} phenomenon manifests most prominently during solar alignment scenarios, where intense glare significantly degrades the visibility of lane demarcations, particularly affecting white and dashed line recognition. 
This optical interference is further amplified under post-precipitation conditions, as aqueous road surfaces generate specular reflections that completely obscure lane markings in visual sensor data.
Analogously, nocturnal \texttt{Streetlight Reflection} substantially impairs lane marking detection accuracy. 
To enhance the realism and severity of this illusion, we incorporate two predominant streetlight chromaticities, \ie, white and yellow, which exhibit spectral congruence with conventional lane marking colors. 
Furthermore, we propose \texttt{Vehicle Reflection} to characterize retroreflective phenomena from vehicle surfaces, which may occlude preceding vehicles in visual field.







\begin{figure}
    \centering
    \includegraphics[width=0.93\linewidth]{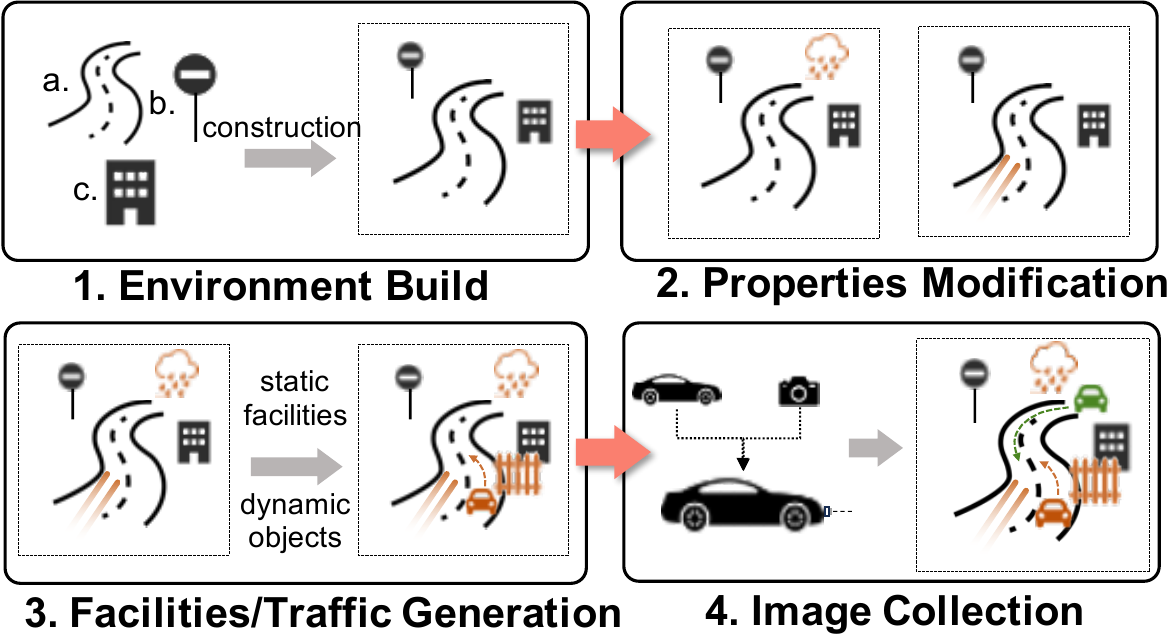}
    \caption{Illustration of our scenario construction pipeline.}
    \label{fig:data_collection}
\end{figure}

\subsection{Scenario Construction}
\label{sec:Construction}

\begin{figure*}[!t]
    \centering
    \includegraphics[width=0.95\linewidth]{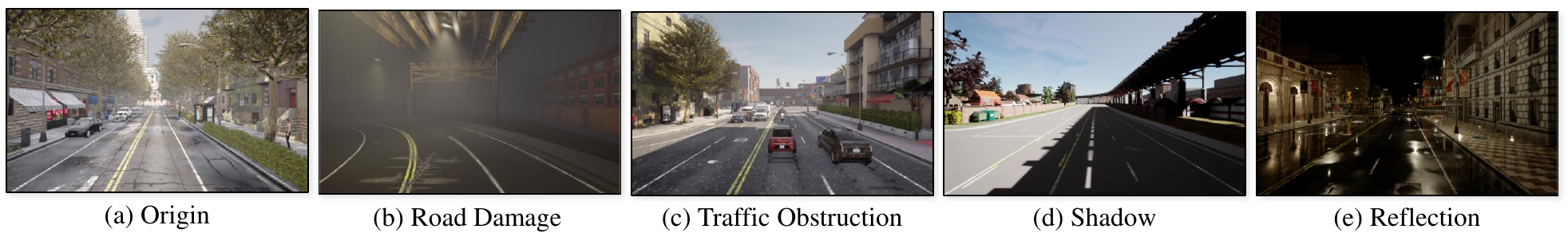}
    \caption{Visualization of images from our \tool dataset under different environmental illusions.}
    \label{fig:dataset_image}
\end{figure*}

While the environmental illusions we examine occur naturally in real-world scenarios, acquiring a comprehensive dataset with sufficient diversity and scale of these phenomena presents significant practical challenges.
To address this limitation, we leverage the CARLA simulator \cite{dosovitskiy2017carla} to systematically generate high-fidelity perturbed scenarios, from which we subsequently extract visually authentic image samples.
Our data collection pipeline operates through two sequential phases: \ding{182} employing real-world traffic patterns as reference, we architect customized 3D environments incorporating prototypical road geometries (including map segmentations), such as T-junctions and other prevalent configurations; \ding{183} implementing the illusion synthesis techniques established in \Sref{sec:corruption_types}, we systematically alter object attributes within the simulated environment;
\ding{184} we strategically position objects and simulate diverse traffic patterns; 
\ding{185} deploying our autonomous agent along predetermined routes, we record each scenario and capture egocentric visual data. The RGB sensor, configured with a resolution of $1280 \times 720$ pixels and a $90.0^{\circ}$ field of view, is mounted at the frontal position of the vehicle agent.
To enhance practical relevance, we adopt the methodology established in \cite{bezzina2014safety_natural_statis,feng2021intelligent_fengshuo} and faithfully reproduce fundamental yet frequently encountered traffic scenarios for autonomous vehicles, 
including car-following maneuvers and sharp turning operations. 
The complete data generation framework and representative captured images are illustrated in \Fref{fig:data_collection} and \Fref{fig:dataset_image}, respectively.

\textbf{Quality control.} 
We adhere to rigorous annotation quality control protocols consistent with established benchmark datasets \cite{tusimple,lin2014microsoft_coco}. 
Our annotation pipeline requires all annotators to strictly follow standardized guidelines specifying both lane annotation targets and methodology.
To ensure annotation fidelity, we implement a three-tier verification process: 
each image undergoes independent annotation by two distinct groups, with the third group performing consensus validation on the averaged results.


\begin{figure}[!t]
    \centering
    \subfloat[]{\includegraphics[width=0.48\linewidth]{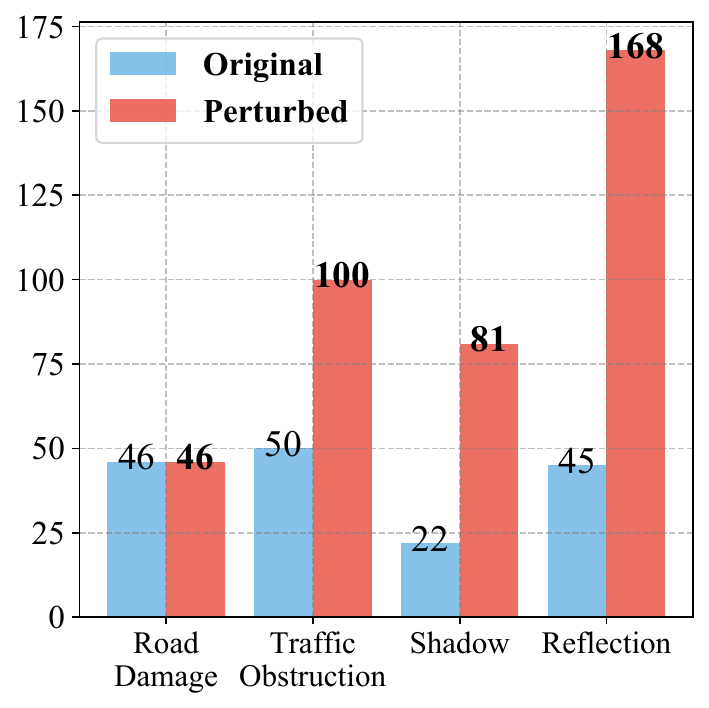}
        \label{fig:dataset_a}
    }
    \subfloat[]{\includegraphics[width=0.48\linewidth]{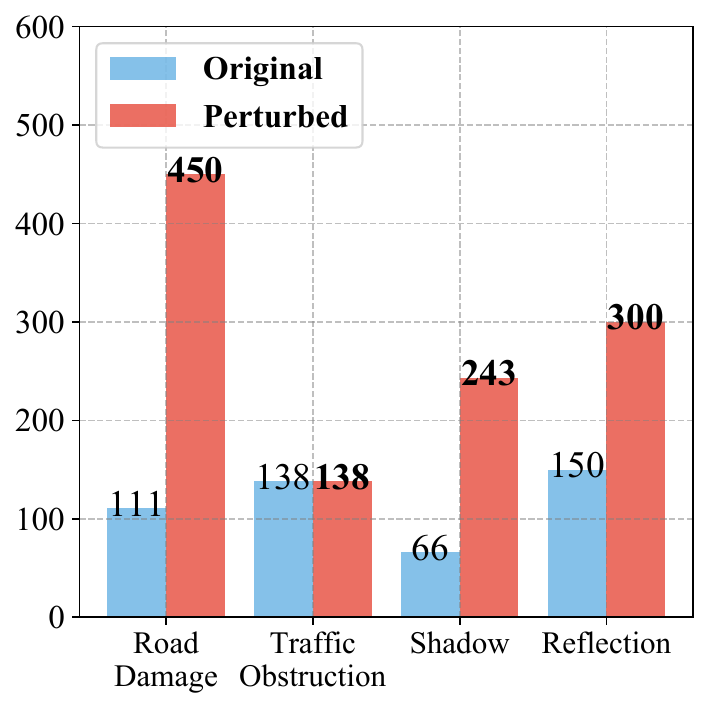}
        \label{fig:dataset_c}
    }
    
    \subfloat[]{
        \includegraphics[width=0.44\linewidth]{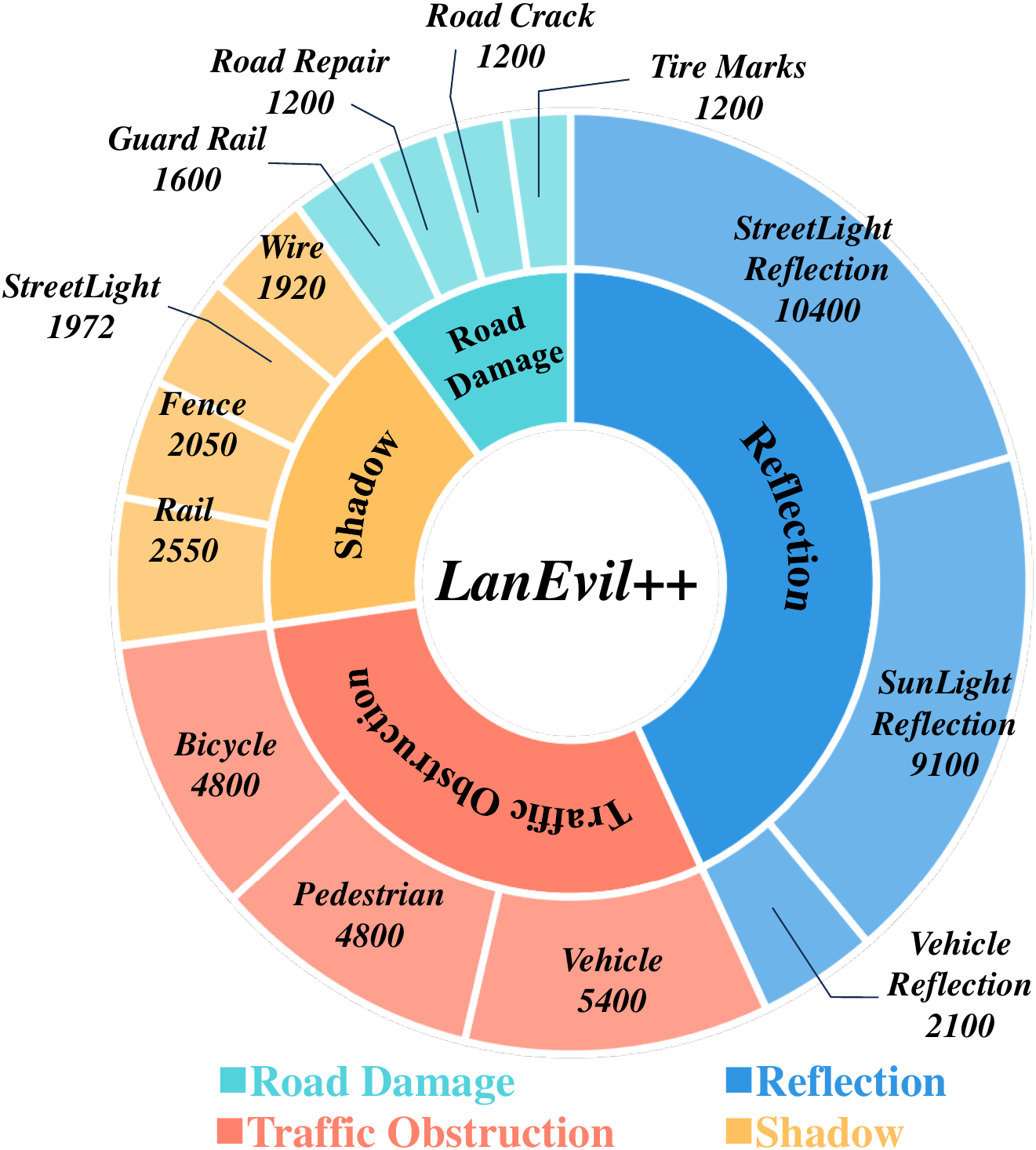}
        \label{fig:dataset_b}
    }
    \subfloat[]{
        \includegraphics[width=0.53\linewidth]{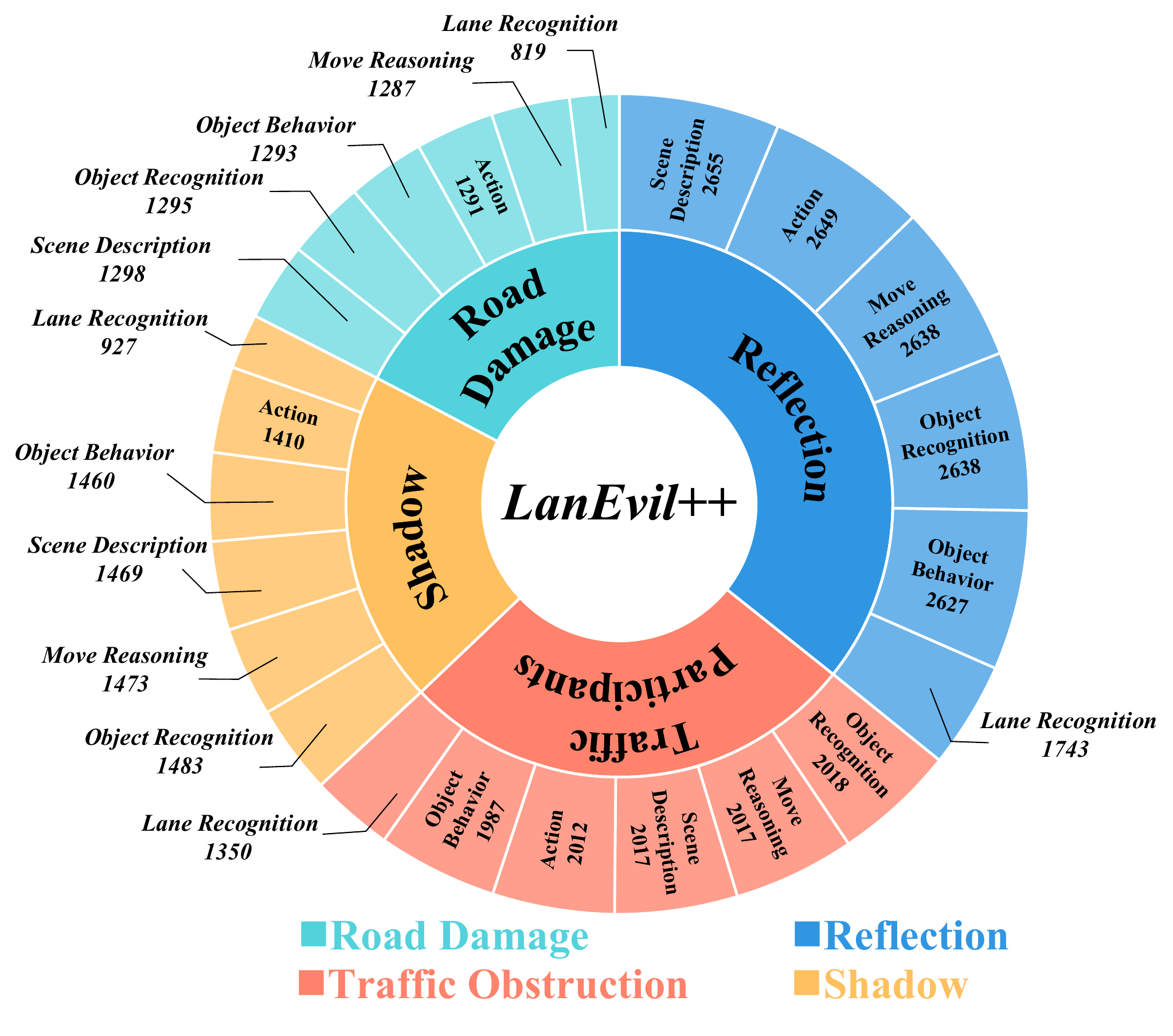}
        \label{fig:dataset_d}
    }

    \caption{The statistics of \tool dataset. (a) The number of original and perturbed cases under four categories. (b) The number of original and perturbed clips under four categories. (c) The case distribution of four illusion categories. (d) The QA pair distribution of four illusion categories.}
    \label{fig:dataset}
\end{figure}

\begin{table}[]
\centering
\caption{Detailed data properties of \toolns.}
\label{tab:diversity}

\renewcommand\arraystretch{1.2}
\begin{tabular}{@{}c@{}}
\subfloat[Scenario diversity]{
\label{tab:diversity_a}
\tiny
\resizebox{\columnwidth}{!}{%
\begin{tabular}{cclllc}
\Xhline{0.6px}
\textbf{Type} & \textbf{Number} & \multicolumn{4}{c}{\textbf{Typical examples}} \\ \hline
Scene & 5 & \multicolumn{4}{c}{Urban, Highway} \\ 
Lane line & 9 & \multicolumn{4}{c}{White single solid, Yellow double solid} \\ 
Weather & 12 & \multicolumn{4}{c}{SoftRainNoon, ClearNoon} \\ 
Road type & 9 & \multicolumn{4}{c}{T-junction, Uphill} \\ 
\Xhline{0.6px}
\end{tabular}%
}} \end{tabular}

\renewcommand\arraystretch{1.2}

\begin{tabular}{@{}c@{}}
\subfloat[Quality distribution]{
\label{tab:diversity_b}
\large
\resizebox{\columnwidth}{!}{%
\begin{tabular}{c|cccc}
\toprule[1.3pt]
\textbf{Image Type} & \textbf{Road Damage} & \textbf{Traffic Obstruction} & \textbf{Shadow} & \textbf{Reflection} \\ \hline
Original & 2,600 & 5,000 & 2,051 & 4,600 \\ 
Perturbed & 2,600 & 10,000 & 6,441 & 17,000 \\ \hline
Total & 5,200 & 15,000 & 8,492 & 21,600 \\
\bottomrule[1.3pt]
\end{tabular}%
}}\end{tabular}
\vspace{-0.1in}
\end{table}

\subsection{Data Generation}

To accommodate the characteristics of \advlmns, \tool extends the original benchmark by additionally incorporating both video clips and QA pairs. Each clip is paired with multiple QA items, carefully designed according to the type of environmental illusion present. This section details the generation process of these two components.

\textbf{Video clip generation.}
In the scenario construction process outlined in \Sref{sec:Construction}, we generate coarse-grained scene videos for each type of environmental illusion, encompassing the entire illusion-affected road segment. 
We perform a refined extraction of the raw videos to create a dataset better suited for model inference.
Specifically, each scene video is segmented into three distinct clips, each consisting of a fixed six-frame sequence. 
The selection of frame rate and segmentation points is determined based on the specific characteristics of the illusion type. 
This refinement process aims to further delineate and emphasize the deceptive nature of each illusion while maximizing coverage across different sampling perspectives and variations in illusion presentation, thereby enhancing the dataset’s diversity and representational richness.
For instance, in the case of the road crack illusion, our clips are segmented to capture the crack at different stages of vehicle interaction, 
specifically before the vehicle passes over it, during the vehicle’s traversal, and as the vehicle moves alongside it.

\begin{figure}[t]
    \centering
    \includegraphics[width=0.95\linewidth]{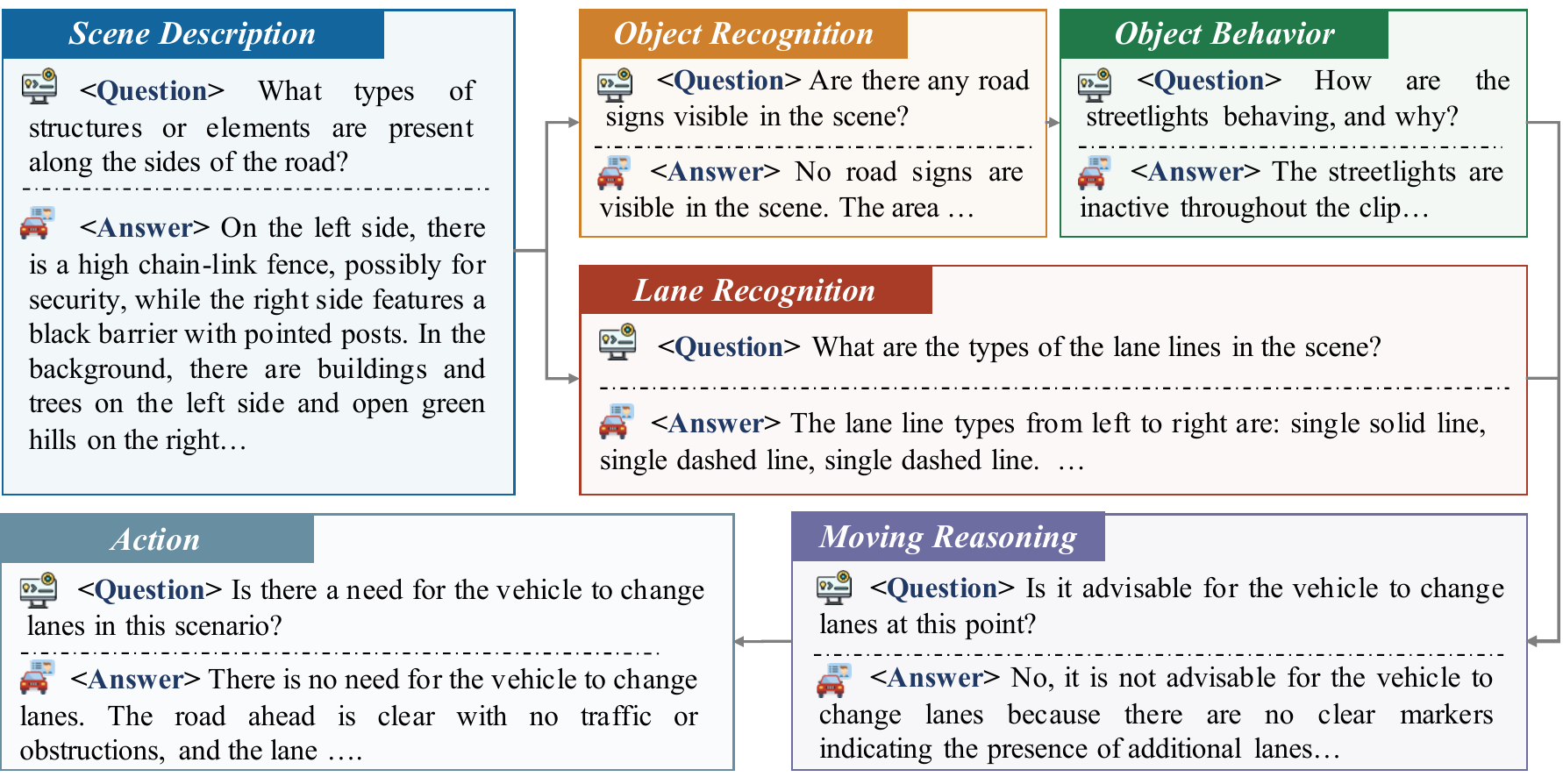}
    \caption{Structure and examples of the QA pairs in \toolns.}
    \label{fig:QA-structure}
\end{figure}

\textbf{Question-answering design.} \tool provides 6 categories of question-answering pairs that are important for AD perception, prediction and planning, including: 
\ding{182} \textbf{Scene Description}, which requires a high-level understanding of the driving environment, such as road conditions, weather, and lighting.
\ding{183} \textbf{Object Recognition}, which focuses on identifying key traffic participants such as vehicles, pedestrians, and traffic signs.
\ding{184} \textbf{Lane Recognition}, which assesses the understanding of lane structures, markings, and boundaries.
\ding{185} \textbf{Object Behavior}, which is related to the movement and interactions of dynamic objects in the scene.
\ding{186} \textbf{Motion Reasoning}, which emphasizes the interpretation and analysis of object behaviors.
\ding{187} \textbf{Action}, which evaluates appropriate driving decisions based on situational awareness and traffic dynamics.
Our QA design adheres to the principle of incorporating illusion-induced cues within the questions, ensuring that each question considers potential deceptive perspectives that may mislead the AD system’s understanding under the influence of the given illusion. The structure and QA examples are shown in \Fref{fig:QA-structure}.

\textbf{Question-answering generation.} 
We leverage the advanced vision-language model GPT-4o \cite{chatgpt} to annotate our dataset. 
Specifically, following the aforementioned design principles, we provide GPT-4o with structured templates to guide the QA generation process, enabling it to generate five QA pairs for each video clip. 
Notably, the same set of QA pairs is shared between the clean dataset and its perturbed counterpart to facilitate direct comparative analysis. 
We conduct a two-turn GPT generation process, during which human experts evaluate and refine the quality of the generated QA pairs. 
Through rigorous assessment and correction, the most optimal QA pairs are selected for inclusion in the final dataset.
\textit{The details of QA generation are in the Supplementary Materials.}

\subsection{Data Properties}
\label{sec:data_property}

\quad \textbf{Subset division.} 
\ding{182} For conventional LD models, 
our \tool comprises two distinct subsets: a training set containing unperturbed reference images and a test set incorporating systematically generated environmental illusions. The training corpus consists of 40,000 randomly selected clean images devoid of artificial perturbations. 
For evaluation purposes, we construct a comprehensive test set of 50,292 images featuring controlled illusion scenarios.

For every fundamental environmental illusion type, we include an unperturbed baseline scenario alongside 1 - 10 systematically modified variants, with each case comprising 50 to 300 temporally coherent driving frames. 
The quantitative distribution of baseline and perturbed scenarios across prevalent lane detection models is presented in \Fref{fig:dataset_a} and \Tref{tab:diversity_b}. 
\ding{183} For \advlmns, 
\tool provides only a test set, which is divided into two subsets: an original scenario set and a perturbed scenario set, containing 465 clips and 1,131 clips, respectively. The statistics of different categories for \advlm are shown in \Fref{fig:dataset_c}.

\textbf{Category distribution.} 
\ding{182} For conventional LD models, the \tool test set comprises 50,292 annotated images spanning 94 fundamental driving scenarios (\eg, straight paths, turning maneuvers) with fully customizable 3D environments. 
\Fref{fig:dataset_b} presents the quantitative distribution of cases and images across each environmental illusion category. 
Furthermore, the \tool benchmark incorporates 9 distinct lane marking types with varying geometries and chromatic characteristics, while encompassing diverse driving contexts and roadway topologies, as demonstrated in \Tref{tab:diversity_a}.
\ding{183} For \advlmns, 
our \tool includes 48,155 QA pairs based on 1,596 fundamental clips. The number of QA pairs for each task is summarized in \Fref{fig:dataset_d}.

\begin{table}[t]
\centering
\caption{\revised{Coverage of 4 environmental illusion categories across representative LD benchmarks. 
Symbols denote coverage status. (\CheckmarkBold: included, {\large$\circ$\normalsize}: partially covered, \XSolidBrush: missing.})}

\label{tab:illusion_coverage_matrix}
\vspace{2mm}
\resizebox{\linewidth}{!}{
\begin{tabular}{c|cccc}
\toprule
\textbf{Dataset} & 
\textbf{Traffic Obstruction} & 
\textbf{Road Damage} & 
\textbf{Shadow} & 
\textbf{Reflection} \\
\midrule
CalTech~\cite{aly2008real} & \large$\circ$ & \XSolidBrush & \large$\circ$ & \large$\circ$ \\
VPGNet~\cite{lee2017vpgnet} & \large$\circ$ & \XSolidBrush & \XSolidBrush & \XSolidBrush \\
TuSimple~\cite{tusimple} & \large$\circ$ & \XSolidBrush & \XSolidBrush & \XSolidBrush \\
CULane~\cite{pan2018spatial_SCNN} & \large$\circ$ & \XSolidBrush & \large$\circ$ & \large$\circ$ \\
LLAMAS~\cite{behrendt2019unsupervised_llamas} & \large$\circ$ & \XSolidBrush & \XSolidBrush & \XSolidBrush \\
BDD100K~\cite{yu2020bdd100k} & \CheckmarkBold & \XSolidBrush & \XSolidBrush & \XSolidBrush \\
CurveLane~\cite{xu2020curvelane} & \large$\circ$ & \XSolidBrush & \XSolidBrush & \XSolidBrush \\
VIL-100 & \CheckmarkBold & \large$\circ$ & \large$\circ$ & \large$\circ$ \\
LanEvil~\cite{zhang2024lanevil} & \CheckmarkBold & \CheckmarkBold & \CheckmarkBold & \CheckmarkBold \\
\rowcolor[HTML]{EFEFEF}
LanEvil++ (Ours) & \CheckmarkBold & \CheckmarkBold & \CheckmarkBold & \CheckmarkBold \\
\bottomrule
\end{tabular}}
\vspace{-0.1in}
\end{table}

\revised{
\textbf{Comparison with existing benchmarks.} 
\Tref{tab:illusion_coverage_matrix} summarizes the coverage of 4 environmental illusion categories across representative LD datasets. Most benchmarks provide only partial coverage of the traffic obstruction illusion, while road damage, shadow, and reflection are largely missing. In contrast, \tool comprehensively encompasses all 4 illusion categories, enabling a more comprehensive evaluation of model robustness to environmental illusions.
}

\textbf{Other application possibilities.} 
Beyond comprehensive image collection, our dataset incorporates rigorously designed large-scale 3D simulation environments with systematic perturbations. These scenarios are preserved in modular file formats to facilitate extensible development, while their dynamic nature enables seamless integration with third-party evaluation systems through CARLA's established interoperability framework.

\section{Multimodal Illusion Defense Approach}\label{sec:CFF}

To tackle environmental illusions, we propose the Multimodal Illusion Defense Approach (MIDA).

Our previous work \cite{zhang2024lanevil} introduced the \textit{Attention Area Mixing} (AAM) method, which focused solely on the visual modality and lacked support for textual inputs, rendering it inadequate for the broader requirements of \advlmns. In this paper, we propose two improved components to address these limitations: \ding{182} AAM++, which incorporates features from both high- and low-attention areas to mitigate bias and further strengthen the model’s resistance to visual distractions; \ding{183} PAT++, a prompt-based adversarial tuning approach specifically designed to enhance robustness against textual perturbations.


\subsection{MIDA for LD Models: Attention Area Mixing++}

\begin{figure}[t]
    \centering
    \includegraphics[width=0.90\linewidth]{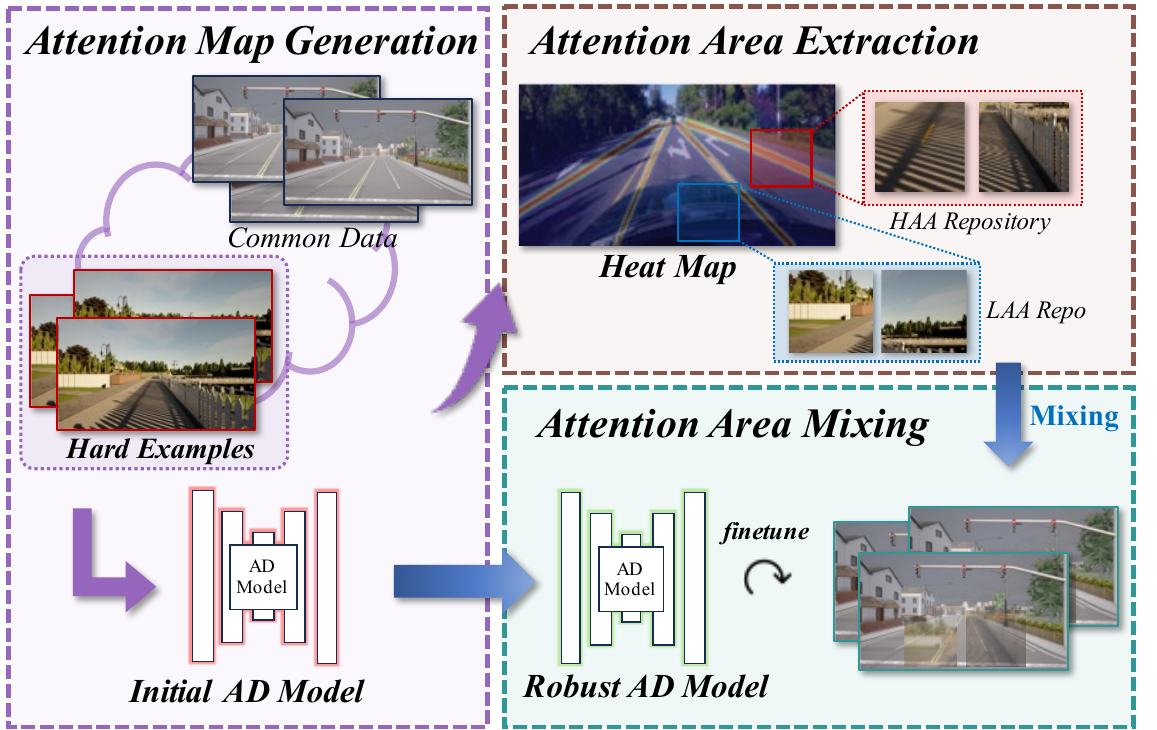}
    \caption{Visual modality of MIDA: Attention Area Mixing++ (AAM++). AAM++ enhances robustness by extracting high- and low-attention regions from hard examples and mixing them separately into training examples.}
    \label{fig:AAM++}
\end{figure}

We first extract high/low attention areas (HAA/LAA) to identify mismatches between the model’s attention and the ground truth. The process begins by applying Gaussian blurring to the heatmap \(M\) to improve generalization:

\begin{equation}
M' = \mathcal{G}(M, \sigma),
\end{equation}

\noindent where \(\mathcal{G}\) denotes the Gaussian blurring function, \(M\) is the attention map generated by an attention module \(\mathcal{H}\) for a given image \(I\) (\ie, \(M = \mathcal{H}(I)\)), and \(M'\) is the resulting blurred heatmap with \(\sigma\) as the standard deviation of the Gaussian kernel. \textit{Details of \(\mathcal{H}\) are provided in the Supplementary Materials.} The method focuses on hard examples, defined as images with accuracy below the dataset mean. Thresholding \(M'\) yields a binary mask \(B\):

\begin{equation}  
B_{HAA}(x, y) =  
\begin{cases}  
1 & \text{if } M'(x, y) > T_{HAA}, \\  
0 & \text{otherwise},  
\end{cases}  
\end{equation}

\noindent where \(T_{HAA}\) is the threshold, and \(M'(x, y)\) represents the value of the blurred attention map at pixel location \((x, y)\). Similarly, the low attention area mask \(B_{LAA}\) is computed using a threshold \(T_{LAA}\) with the condition \(M'(x, y) \leq T_{LAA}\). Connected components in \(B\) smaller than a size threshold \(S\) are discarded, resulting in the final interest regions \(\mathbb{A}_k\), where \(k \in \{\text{HAA}, \text{LAA}\}\).

\begin{equation}  
\mathbb{A}_k = \{r \mid \text{area}(r) > S, r \subset B_k\}.  
\end{equation}  

Discrepancies are identified by comparing each region \(r \in \mathbb{A}_k\) with the ground truth \(G\), defining mismatches as regions with insufficient overlap and extracting the minimum bounding rectangles (MBR) for these regions:

\begin{equation}
\mathbb{R}_{k} = \{\text{MBR}(r) \mid r \in \mathbb{A}_{k} \wedge \text{overlap}(r, G) < \theta\},
\end{equation}
where $\mathbb{R}_{k}$ represents the set of mismatched regions enclosed by minimum bounding rectangles, and \(\theta\) establishes the threshold for acceptable overlap. After generating \(\mathbb{R}_{k}\), we mix hard examples into the dataset. For each image \(I\) in \(\mathbb{D}\), we integrate a randomly selected attention area \(H_{k}\) from \(\mathbb{R}_{k}\):
 
\begin{equation}  
I^{\text{mixed}}_{k} = I \oplus \mathcal{S}_{k}(H_{k}, G_I),  
\end{equation}  
where \(I^{\text{mixed}}_{k}\) is the augmented image incorporating \(H_{k}\) based on the ground truth \(G_I\), and \(\oplus\) represents the blending operation. The \(\mathcal{S}_{k}\) function searches for low or high attention areas in the ground truth, opposite to the type of \(k\). That is, \(\mathcal{S}_{HAA}\) locates low attention areas. The framework of AAM++ is illustrated in \Fref{fig:AAM++}.

\subsection{MIDA for \advlmns: Prompt Adversarial Tuning++}

\begin{figure}[t]
    \centering
    \includegraphics[width=0.90\linewidth]{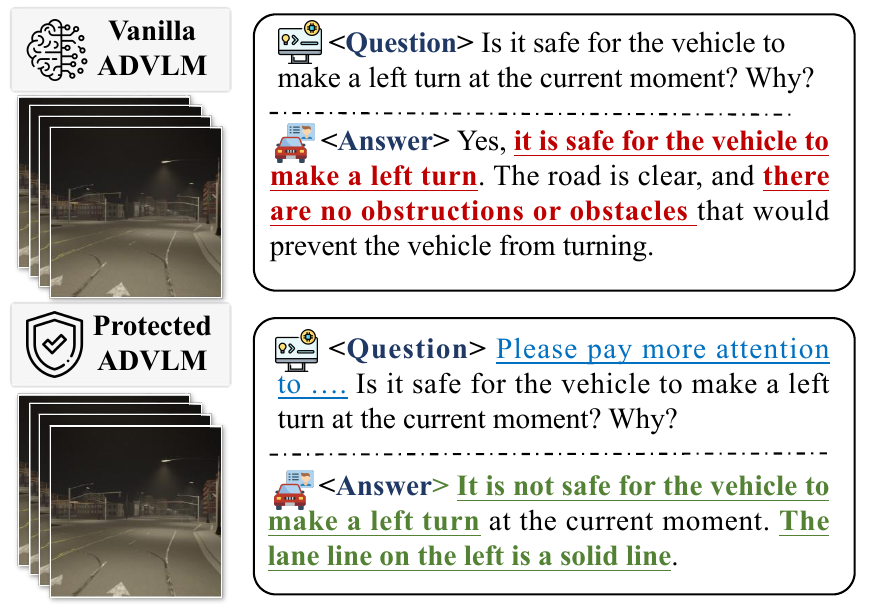}
    \caption{Textual modality of MIDA: Prompt Adversarial Tuning++ (PAT++). PAT++ enhances robustness against environmental illusions through prefix-based prompt tuning. Illustrated using DriveLM, where the ground illusion \texttt{Tiremarks} misleads lane recognition. With the addition of a protective prefix (shown in blue), DriveLM correctly identifies the lane markings.}
    \label{fig:PAT++}
    \vspace{-0.1in}
\end{figure}

The textual modality, with its discontinuous nature and large search space, poses unique challenges for \advlmns and does not lend itself to visual mixing-based approaches. While the original PAT \cite{mo2024fight} was devised for LLMs, it did not account for image interactions. Therefore, we enhance PAT with a specialized prompt-tuning scheme, bolstering the model’s robustness against text-based adversarial illusions in multimodal contexts.

Given a visual input \(\mathbb{I} \subset \mathbb{I}_\text{all}\) and a textual input \(t \in \mathbb{T}\), an ADVLM generates a response \(\mathbf{res} = t_{n+1:n+L}\), where each \(t_n\) is a token. The likelihood of generating the next token sequence is defined as:

\begin{equation}  
p(t_{n+1:n+L} | \mathbb{I}, t) = \prod_{i=1}^{L} p(t_{n+i} | t_{1:n+i-1}, \mathbb{I}, t).  
\end{equation}  

The training objective is to minimize the negative log-likelihood of generating the target sequence \(t_{n+1:n+L}\) given the visual and textual inputs:

\begin{equation}  
\mathcal{L}(\mathbb{I}, t) = -\log p(t_{n+1:n+L} | \mathbb{I}, t_{n}).  
\end{equation}

Let \(\tilde{t}_{1:n} = \{\tilde{t}_{1}, \tilde{t}_{2}, \ldots, \tilde{t}_{n}\}\) denote the defense prompt, where \(\tilde{t}_{i} \in \mathbb{T}\) represents the \(i\)-th token of the defense prompt. Then the textual prompt can be defined as \(\text{CONCAT}(\tilde{t}_{1:n}, t)\), where \(\text{CONCAT}(\cdot)\) denotes the concatenation operation over the token sequences.  

Similar to the visual modality, we distinguish between benign examples \(\mathbb{I}_{\text{bgn}}\) and hard examples \(\mathbb{I}_{\text{hard}}\). For hard examples, we aim to enhance the model’s focus on the illusion area by adjusting the defense prompt. To achieve this, we define a malicious target \(\mathbf{res}_{goal}\) (\ie, ``I will focus on the factors affecting the road''). The loss function for hard examples is then formulated as:

\begin{equation}  
\mathcal{L}_{\text{hard}}(\tilde{t}_{1:n}, \mathbb{I}_{\text{hard}}, \mathbf{res}_{goal}) = -\log p(\mathbf{res}_{goal} | \mathbb{I}_{\text{hard}}, \tilde{t}_{1:n}).  
\end{equation}  

For benign examples, the goal is to enable the model to respond normally without interference from the defense prompt. The loss function for benign examples is defined as:  

\begin{equation}  
\mathcal{L}_{\text{bgn}}(t_{1:n}, \mathbb{I}_{\text{bgn}}, \mathbf{res}_{gt}) = -\log p(\mathbf{res}_{gt} | \mathbb{I}_{\text{bgn}}, t_{1:n}).  
\end{equation}  

Combining the two stages, the general optimization objective can be formulated as:

\begin{equation}  
\tilde{t}^{\star} = \arg\min_{\tilde{t} \in \mathbb{T}}  
\begin{aligned}  
 \alpha & \mathcal{L}_{\text{bgn}}(t_{1:n}, \mathbb{I}_{\text{bgn}}, \mathbf{res}_{gt}) + \\  
 (1 - \alpha) &\mathcal{L}_{\text{hard}}(\tilde{t}_{1:n}, \mathbb{I}_{\text{hard}}, \mathbf{res}_{goal}),  
\end{aligned}  
\end{equation}  
where \(\alpha \in [0, 1]\) is a hyper-parameter that balances the loss terms. This adjustment ensures that the defense prompt enhances the model’s robustness to hard examples while maintaining normal responses for benign cases.
\section{Experiments}
\label{sec:experiments}

\begin{table*}
\centering
\caption{Evaluation results of different LD models using ResNet-18 on the \tool dataset. LD models are trained using the \tool training set. For each category of illusion, we report the average value over different types and severity levels. The \textbf{bold} values represent the maximum in each column, and ``Gap'' is computed by ``Perturbed'' minus ``Original''.}
\label{tab:main_results}

\renewcommand\arraystretch{1.2}

\begin{tabular}{@{}c@{}}
\label{tab:main_results_ACC}
\subfloat[Results under Accuracy (\%)]{
\huge
\resizebox{0.99\textwidth}{!}{%
\begin{tabular}{c|cc
>{\columncolor[HTML]{EFEFEF}}c|cc
>{\columncolor[HTML]{EFEFEF}}c|cc
>{\columncolor[HTML]{EFEFEF}}c|cc
>{\columncolor[HTML]{EFEFEF}}c|cc 
>{\columncolor[HTML]{EFEFEF}}c }
\toprule[2.0pt]
 & \multicolumn{3}{c|}{Road Damage} & \multicolumn{3}{c|}{Traffic Obstruction} & \multicolumn{3}{c|}{Shadow} & \multicolumn{3}{c|}{Reflection} & \multicolumn{3}{c}{\emph{Average}} \\ \cmidrule(l){2-16} 
\multirow{-2}{*}{\textbf{Method}}  & Perturbed & Original & Gap & Perturbed & Original & Gap & Perturbed & Original & Gap & Perturbed & Original & Gap & Perturbed & Original & Gap \\ \midrule
LaneATT \cite{tabelini2021keep_LaneATT} & 76.23 & 78.66 & -2.43 & 73.18 & 75.84 & -2.66 & 79.48 & 86.07 & -6.59 & 71.33 & 78.93 & -7.59 & 73.53 & 78.85 & -5.32 \\ 
UFLD \cite{qin2020ultra} & 65.90 & 68.47 & -2.57 & 65.64 & 67.58 & -1.94 & 67.73 & 74.01 & -6.27 & 64.45 & 70.88 & -6.42 & 65.45 & 69.90 & -4.44 \\ 
BezierLaneNet \cite{feng2022rethinking} & 73.14 & 75.76 & -2.62 & 71.44 & 74.51 & -3.07 & 68.79 & 78.32 & -9.54 & 69.44 & 75.31 & -5.87 & 70.00 & 75.52 & -5.52 \\ 
GANet \cite{wang2022keypoint_GANet} & 85.32 & 89.33 & \textbf{-4.01} & 80.53 & 84.75 & \textbf{-4.22} & 83.44 & 93.02 & \textbf{-9.58} & 79.21 & 89.23 & \textbf{-10.02} & 80.55 & 88.08 & \textbf{-7.53} \\ 
SCNN \cite{pan2018spatial_SCNN} & 71.11 & 71.94 & -0.84 & 67.17 & 69.67 & -2.50 & 65.38 & 70.33 & -4.95 & 63.88 & 68.05 & -4.16 & 65.31 & 69.36 & -4.05 \\
\revised{LaneFormer \cite{han2022laneformer}} & \revised{87.14} & \revised{90.06} & \revised{-2.92} & 
\revised{83.72} & \revised{87.25} & \revised{-3.53} & 
\revised{86.20} & \revised{93.41} & \revised{-7.21} & 
\revised{81.68} & \revised{89.20} & \revised{-7.52} & 
\revised{84.69} & \revised{89.98} & \revised{-5.29} \\

\revised{CLRNet \cite{zheng2022clrnet}} & \revised{86.02} & \revised{88.77} & \revised{-2.75} & 
\revised{82.63} & \revised{85.44} & \revised{-2.81} & 
\revised{84.95} & \revised{91.22} & \revised{-6.27} & 
\revised{80.37} & \revised{87.42} & \revised{-7.05} & 
\revised{83.49} & \revised{88.21} & \revised{-4.72} \\

\bottomrule[2.0pt]
\end{tabular}%
}}\end{tabular}

\renewcommand\arraystretch{1.2}
\begin{tabular}{@{}c@{}}
\label{tab:main_results_F1}
\subfloat[Results under F1-score (\%)]{
\huge
\resizebox{0.99\textwidth}{!}{%
\begin{tabular}{c|cc
>{\columncolor[HTML]{EFEFEF}}c|cc
>{\columncolor[HTML]{EFEFEF}}c|cc
>{\columncolor[HTML]{EFEFEF}}c|cc
>{\columncolor[HTML]{EFEFEF}}c|cc 
>{\columncolor[HTML]{EFEFEF}}c }
\toprule[2.0pt]
 & \multicolumn{3}{c|}{Road Damage} & \multicolumn{3}{c|}{Traffic Obstruction} & \multicolumn{3}{c|}{Shadow} & \multicolumn{3}{c|}{Reflection} & \multicolumn{3}{c}{\emph{Average}} \\ \cmidrule(l){2-16} 
\multirow{-2}{*}{\textbf{Method}}  & Perturbed & Original & Gap & Perturbed & Original & Gap & Perturbed & Original & Gap & Perturbed & Original & Gap & Perturbed & Original & Gap \\ \midrule
LaneATT \cite{tabelini2021keep_LaneATT} & 49.20 & 52.63 & -3.43 & 46.03 & 51.87 & -5.84 & 48.23 & 60.35 & -12.12 & 38.81 & 53.23 & -14.41 & 42.95 & 53.80 & -10.85 \\ 
UFLD \cite{qin2020ultra} & 21.25 & 24.54 & -3.30 & 28.23 & 31.85 & -3.62 & 26.18 & 36.32 & -10.13 & 19.96 & 30.82 & -10.85 & 23.57 & 31.65 & -8.08 \\ 
BezierLaneNet \cite{feng2022rethinking} & 37.84 & 40.68 & -2.85 & 42.53 & 50.03 & -7.49 & 41.41 & 50.93 & -9.52 & 37.06 & 45.86 & -8.81 & 39.49 & 47.95 & -8.46 \\ 
GANet \cite{wang2022keypoint_GANet} & 79.01 & 84.16 & \textbf{-5.15} & 70.34 & 80.79 & \textbf{-10.45} & 73.50 & 89.33 & \textbf{-15.83} & 65.25 & 83.03 & \textbf{-17.78} & 68.64 & 83.25 & \textbf{-14.61} \\ 
SCNN \cite{pan2018spatial_SCNN} & 48.69 & 52.88 & -4.19 & 43.63 & 50.47 & -6.85 & 34.92 & 49.42 & -14.50 & 34.70 & 46.94 & -12.24 & 37.71 & 49.19 & -11.49 \\ 
\revised{LaneFormer \cite{han2022laneformer}} & \revised{82.12} & \revised{86.57} & \revised{-4.45} & 
\revised{74.65} & \revised{82.91} & \revised{-8.26} & 
\revised{76.44} & \revised{90.24} & \revised{-13.80} & 
\revised{69.52} & \revised{84.71} & \revised{-15.19} & 
\revised{75.68} & \revised{86.11} & \revised{-10.43} \\

\revised{CLRNet \cite{zheng2022clrnet}} & \revised{80.93} & \revised{84.98} & \revised{-4.05} & 
\revised{72.44} & \revised{80.36} & \revised{-7.92} & 
\revised{74.35} & \revised{88.02} & \revised{-13.67} & 
\revised{67.83} & \revised{82.25} & \revised{-14.42} & 
\revised{73.89} & \revised{83.40} & \revised{-9.51} \\

\bottomrule[2.0pt]
\end{tabular}%
}}\end{tabular}
\vspace{-0.1in}
\end{table*}


\subsection{Experimental Setup}

\quad \textbf{Dataset.} For conventional LD models, we first train the LD models from scratch using the training set of our \toolns, and then evaluate their robustness on the test set of \toolns. For \advlmns, we use the pretrained model and fine-tune on the \toolns.
We also evaluate the domain gap between simulated data and real data in \Sref{sec:visual-fid}.

\textbf{Target models.}
\revised{For conventional LD models, we evaluate seven representative methods covering the five LD categories introduced in \Sref{sec:lane_detection_models}: SCNN~\cite{pan2018spatial_SCNN}, GANet~\cite{wang2022keypoint_GANet}, LaneATT~\cite{tabelini2021keep_LaneATT}, CLRNet~\cite{zheng2022clrnet}, LaneFormer~\cite{han2022laneformer}, UFLD~\cite{qin2020ultra}, and BezierLaneNet~\cite{feng2022rethinking}. }
\revised{Among them, CLRNet and LaneFormer represent recent anchor-based architectures enhanced by transformer-based reasoning, enabling more global lane-level understanding.} 
We use ResNet-18/34/50/101 \cite{he2016deep_resnet} with ImageNet-pretrained weights \cite{deng2009imagenet} as backbones. For SCNN, we follow its official implementation and use VGG-16 \cite{simonyan2014very_vgg}. 
\revised{For \advlmns, we use DriveLM \cite{sima2024drivelm}, Dolphins \cite{ma2024dolphins}, Omni-L and Omni-Q \cite{wang2025omnidrive} as baselines, as they are representative open-source models.}

\textbf{Evaluation metrics.}
For conventional LD models, we adopt the two most widely used metrics, \emph{Accuracy} and \emph{F1 score}, to evaluate model performance, both of which are tailored to reflect the spatial precision and detection quality in LD.
\emph{Accuracy} quantifies the proportion of correctly predicted lane points relative to the total number of ground truth points.
A lane point within 20 pixels of the ground truth point is considered correct. 
Lane predictions are considered to be true positives (TP) only when more than 85\% of points are correct.   
\emph{F1 score} serves as a balanced metric by combining Precision and Recall via their harmonic mean, where $Precision = \frac{TP}{TP + FP} $ and $Recall = \frac{TP}{TP + FN}$.
For \advlmns, we use the \emph{GPT score} and \emph{Language score}  as the metric. 
\emph{GPT score} measures the semantic alignment between the generated output and the expected response by the powerful ChatGPT.
\emph{Language score} evaluates the linguistic quality and fluency of the generated text, independent of its semantic correctness, which includes BLEU, ROUGE-L,  METEOR, CIDEr and SPICE.
For all metrics, higher values indicate \emph{better} performance and robustness. 


\subsection{Main Results for Conventional LD models}

We first evaluate the robustness of \revised{seven} conventional LD models on \toolns. Due to space limitations, we report the average performance of models on each of the four main illusion categories here. 
As shown in \Tref{tab:main_results}, we can \textbf{identify}:

\ding{182} Overall, the proposed environmental illusions have demonstrated certain impacts on the robustness of LD models. In general, these illusions can cause an average absolute \revised{\textbf{5.27\%}} Accuracy drop and \revised{\textbf{10.49\%}} F1-Score drop.

\ding{183} Different environmental illusions show different threat impacts on LD model robustness. For instance, \texttt{Shadow} demonstrates the most pronounced effect, leading to an average model performance decrease of \revised{7.20\%}; in contrast, \texttt{Road Damage} shows comparatively weak influence with only \revised{2.59\%} decreases on average.

\ding{184} Following a comprehensive analysis of the model and backbone, we observe that different models display various degrees of resistance to these types of corruption. In particular, GANet showcases the highest clean performance. However, the Accuracy decreases the most after attacks, amounting to approximately 7.53\%. Conversely, SCNN has the lowest Accuracy in clean images but experiences the least Accuracy decrease. Additionally, in terms of model backbones, as the depth of layers increases, the performance and robustness of the model tend to improve.

\revised{\ding{185} Transformer-based architectures such as LaneFormer and CLRNet achieve the highest overall performance, with average clean accuracies of 89.98\% and 88.21\%, respectively, and moderate degradation gaps of 5.29\% and 4.72\%.}

\ding{186} As the severity level increases, the performance degeneration increases significantly. Specifically, the level-1 images cause an average \revised{2.51\%} Accuracy drop and \revised{6.94\%} F1-Score drop; in contrast, the level-5 images can cause an average \revised{19.07\%} Accuracy drop and \revised{33.89\%} F1-Score drop. 

\begin{table*}
\centering
\caption{Evaluation results of different \advlm on the \tool dataset. For each category of illusion, we report the average value over different types and severity levels. The \textbf{bold} values represent the maximum in each column, and ``Gap'' is computed by ``Perturbed'' minus ``Original''.}
\label{tab:main_results_ADVLM}

\renewcommand\arraystretch{1.2}

\begin{tabular}{@{}c@{}}
\label{tab:main_results_gpt}
\subfloat[Results under GPT Score (\%)]{
\huge
\resizebox{0.99\textwidth}{!}{%
\begin{tabular}{c|cc
>{\columncolor[HTML]{EFEFEF}}c|cc
>{\columncolor[HTML]{EFEFEF}}c|cc
>{\columncolor[HTML]{EFEFEF}}c|cc
>{\columncolor[HTML]{EFEFEF}}c|cc 
>{\columncolor[HTML]{EFEFEF}}c }
\toprule[2.0pt]
 & \multicolumn{3}{c|}{Road Damage} & \multicolumn{3}{c|}{Traffic Obstruction} & \multicolumn{3}{c|}{Shadow} & \multicolumn{3}{c|}{Reflection} & \multicolumn{3}{c}{\emph{Average}} \\ \cmidrule(l){2-16} 
\multirow{-2}{*}{\textbf{Method}}  & Perturbed & Original & Gap & Perturbed & Original & Gap & Perturbed & Original & Gap & Perturbed & Original & Gap & Perturbed & Original & Gap \\ \midrule
DriveLM \cite{sima2024drivelm} & 80.45 & 81.61 & -1.16 & 77.78 & 82.01 & \textbf{-4.24} & 84.04 & 84.55 & -0.51 & 79.34 & 82.76 & \textbf{-3.41} & 80.91 & 83.28 & \textbf{-2.38} \\ 
Dolphins \cite{ma2024dolphins} & 62.15 & 63.44 & \textbf{-1.29} & 64.02 & 65.51 & -1.49 & 65.63 & 65.96 & -0.34 & 63.13 & 64.52 & -1.39 & 63.13 & 64.52 & -1.39 \\ 
\revised{Omni-L \cite{wang2025omnidrive}} & \revised{82.91} & \revised{84.02} & \revised{-1.11} & \revised{80.64} & \revised{84.53} & \revised{-3.89} & \revised{85.27} & \revised{86.05} & \revised{\textbf{-0.78}} & \revised{81.46} & \revised{84.72} & \revised{-3.26} & \revised{82.57} & \revised{84.83} & \revised{-2.26} \\ 
\revised{Omni-Q \cite{wang2025omnidrive}} & \revised{84.25} & \revised{85.33} & \revised{-1.08} & \revised{81.72} & \revised{85.14} & \revised{-3.42} & \revised{86.18} & \revised{86.79} & \revised{-0.61} & \revised{82.03} & \revised{85.38} & \revised{-3.35} & \revised{83.55} & \revised{85.66} & \revised{-2.11} \\ 

\bottomrule[2.0pt]
\end{tabular}%
}}\end{tabular}

\renewcommand\arraystretch{1.2}
\begin{tabular}{@{}c@{}}
\label{tab:main_results_language}
\subfloat[Results under Language Score (\%)]{
\huge
\resizebox{0.99\textwidth}{!}{%
\begin{tabular}{c|cc
>{\columncolor[HTML]{EFEFEF}}c|cc
>{\columncolor[HTML]{EFEFEF}}c|cc
>{\columncolor[HTML]{EFEFEF}}c|cc
>{\columncolor[HTML]{EFEFEF}}c|cc 
>{\columncolor[HTML]{EFEFEF}}c }
\toprule[2.0pt]
 & \multicolumn{3}{c|}{Road Damage} & \multicolumn{3}{c|}{Traffic Obstruction} & \multicolumn{3}{c|}{Shadow} & \multicolumn{3}{c|}{Reflection} & \multicolumn{3}{c}{\emph{Average}} \\ \cmidrule(l){2-16} 
\multirow{-2}{*}{\textbf{Method}}  & Perturbed & Original & Gap & Perturbed & Original & Gap & Perturbed & Original & Gap & Perturbed & Original & Gap & Perturbed & Original & Gap \\ \midrule
DriveLM \cite{sima2024drivelm} & 29.53 & 30.21 & \textbf{-0.68} & 28.84 & 30.12 & -1.28 & 32.17 & 32.61 & \textbf{-0.44} & 28.76 & 30.68 & \textbf{-1.92} & 30.54 & 31.43 & \textbf{-0.89}  
 \\ 
Dolphins \cite{ma2024dolphins} & 15.72 & 16.01 & -0.29 & 16.84 & 16.48 & 0.36 & 16.78 & 16.85 & -0.07 & 15.94 & 16.29 & -0.35 & 15.92 & 16.34 & -0.42   \\ 
\revised{Omni-L \cite{wang2025omnidrive}} & \revised{30.46} & \revised{31.02} & \revised{-0.56} & \revised{29.37} & \revised{30.54} & \revised{-1.17} & \revised{33.21} & \revised{33.65} & \textbf{\revised{-0.44}} & \revised{29.86} & \revised{31.08} & \revised{-1.22} & \revised{30.73} & \revised{31.57} & \revised{-0.84} \\ 
\revised{Omni-Q \cite{wang2025omnidrive}} & \revised{31.58} & \revised{32.09} & \revised{-0.51} & \revised{30.12} & \revised{31.42} & \revised{\textbf{-1.30}} & \revised{33.98} & \revised{34.37} & \revised{-0.39} & \revised{30.54} & \revised{31.72} & \revised{-1.18} & \revised{31.56} & \revised{32.40} & \revised{-0.84} \\ 

\bottomrule[2.0pt]
\end{tabular}%
}}\end{tabular}
\vspace{-0.1in}
\end{table*}

\subsection{Main Results for \advlmns}

We then evaluate the robustness of \revised{four} \advlm on \toolns. As shown in \Tref{tab:main_results_ADVLM}, we can \textbf{identify}:

\ding{182} Consistent with observations in conventional LD models, \advlm also exhibit significant vulnerability to environmental illusions, reflected by a \revised{\textbf{2.03\%}} decrease in average GPT score and a \revised{\textbf{0.75\%}} drop in average Language score.

\ding{183} Different categories of environmental illusions have varying impacts. 
\revised{Among them, \texttt{Traffic Obstruction} causes the most severe decline, with an average GPT Score drop of 3.51\% and a Language Score drop of 1.04\%. 
In contrast, \texttt{Shadow} leads to the smallest decline, with average GPT and Language Score decreases of only 0.56\% and 0.33\%, respectively. This differs from conventional LD models, implying distinct attention patterns in \advlmns.} 

\revised{\ding{184} Dolphins show the highest robustness under environmental illusions (GPT Score: 1.39\%, Language Score: 0.42\%), whereas DriveLM exhibits the largest performance degradation (2.38\%, 0.89\%), with Omni-Q (2.11\%, 0.84\%) and Omni-L (2.26\%, 0.84\%) lying in between.}

\ding{185} Similar to conventional LD models, performance degradation in \advlm becomes more severe as the illusion level increases. \revised{Level-1 images lead to an average GPT Score drop of \revised{1.34\%} and a Language Score drop of \revised{0.38\%}, while level-5 images cause much larger average decreases of 9.11\% and 2.62\%, respectively.}

\subsection{Results on Noise Defense Methods}
\label{sec:noise_defense}

In this section, we further investigated the effectiveness of our proposed MIDA method and existing noise defense methods \cite{sun2023improving,liu2023exploring,liang2023exploring,liang2024unlearning} on \tool dataset.

\begin{figure}[!t]
    \centering  
    \subfloat[LaneATT]{%
        \includegraphics[width=0.215\textwidth]{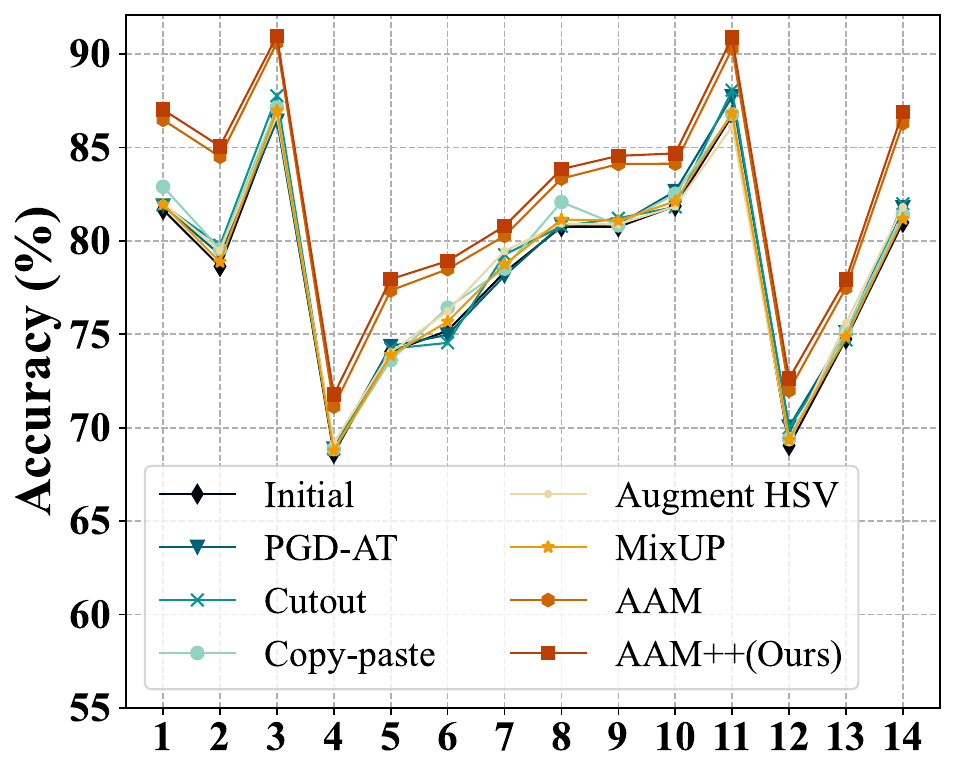}%
        \label{fig:laneatt}%
    }
    \hspace{0.02\textwidth}  
    \subfloat[UFLD]{%
        \includegraphics[width=0.215\textwidth]{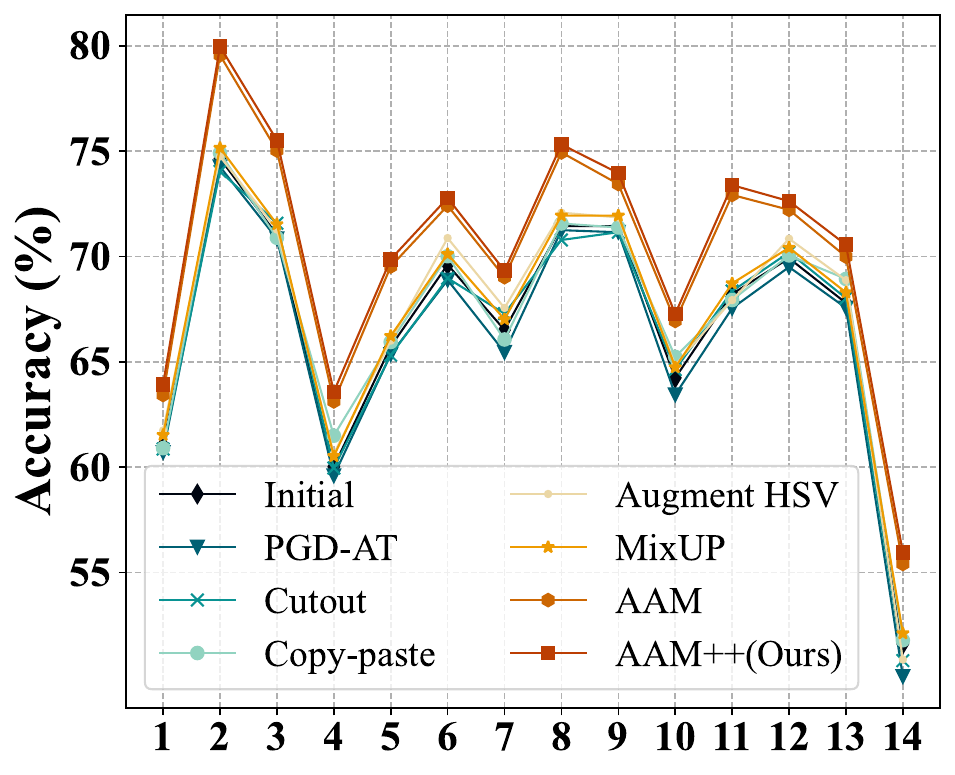}%
        \label{fig:ufld}%
    }

    \subfloat[DriveLM]{%
        \includegraphics[width=0.215\textwidth]{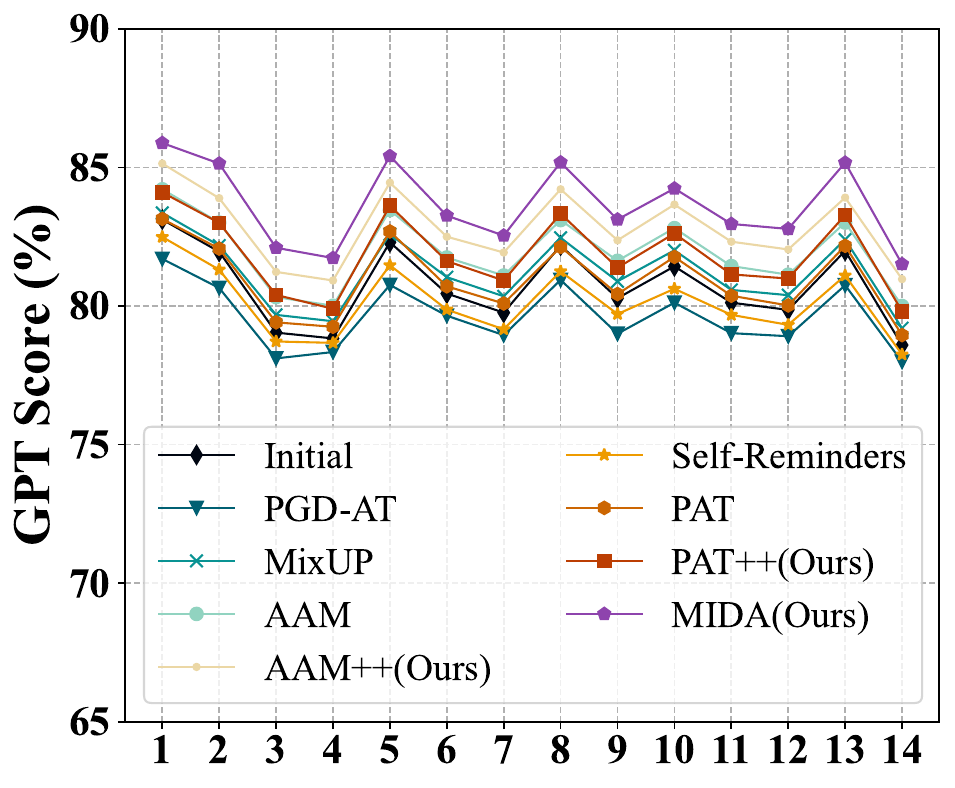}%
        \label{fig:drivelm}%
    }
    \hspace{0.02\textwidth}  
    \subfloat[Dolphins]{%
        \includegraphics[width=0.215\textwidth]{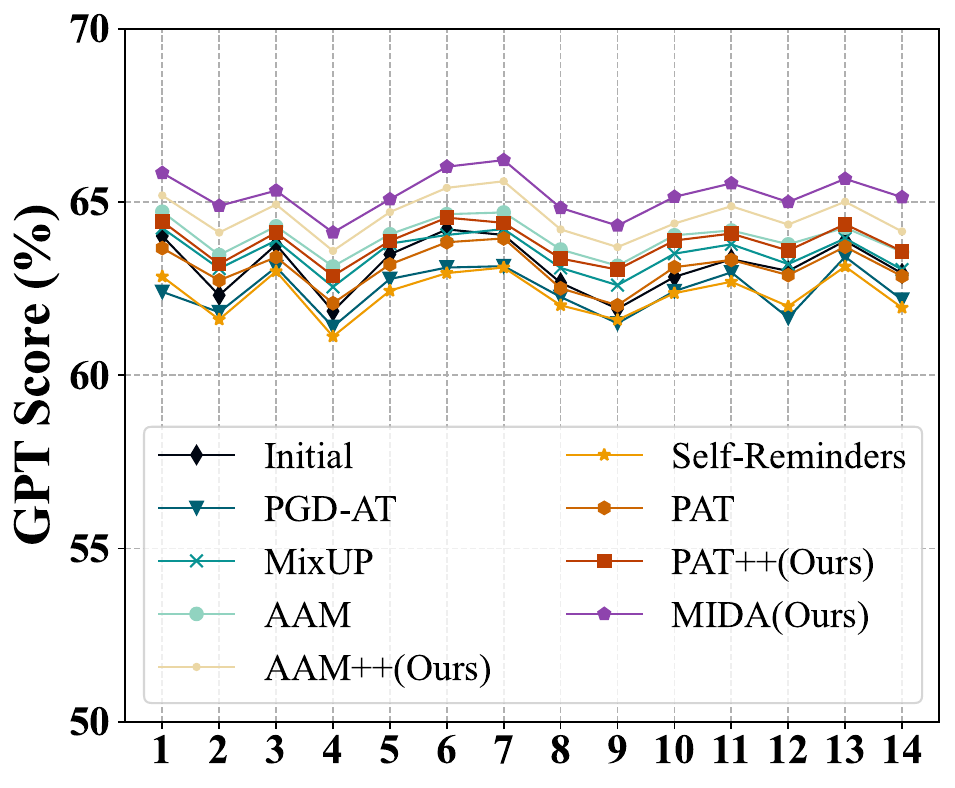}%
        \label{fig:dolphins}%
    }

    \subfloat[\revised{LaneATT$^\dagger$}]{%
        \includegraphics[width=0.215\textwidth]{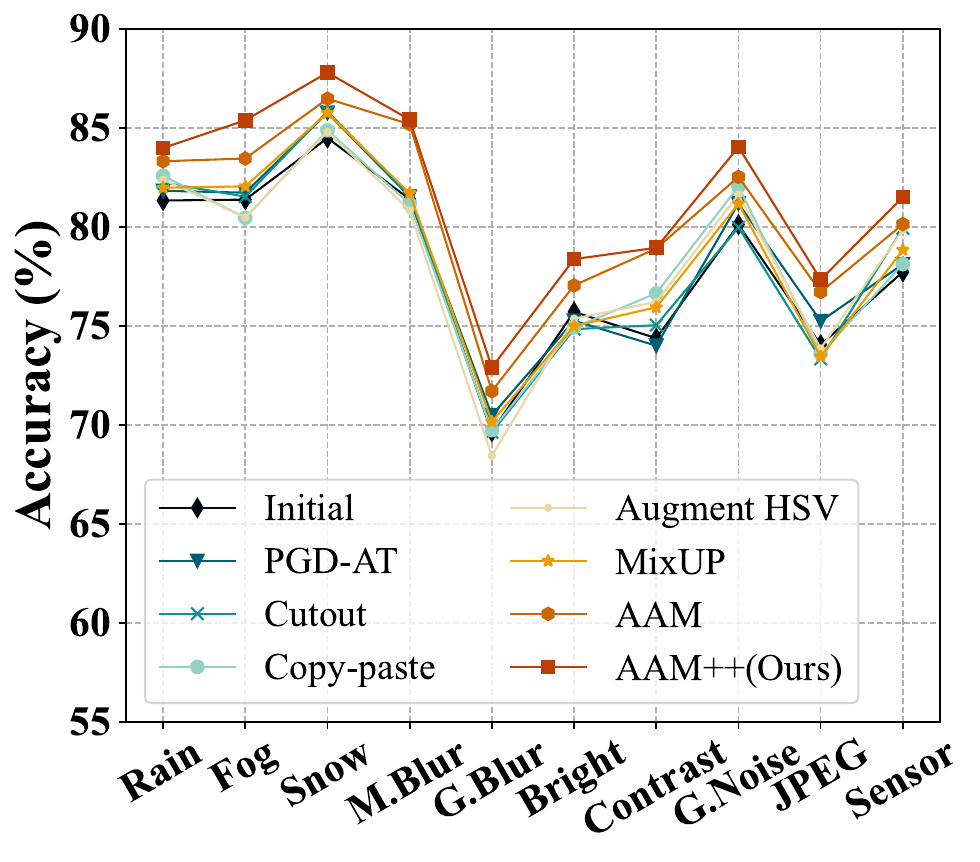}%
        \label{fig:laneatt_nonillusion}%
    }
    \hspace{0.02\textwidth}
    \subfloat[\revised{DriveLM$^\dagger$}]{%
        \includegraphics[width=0.215\textwidth]{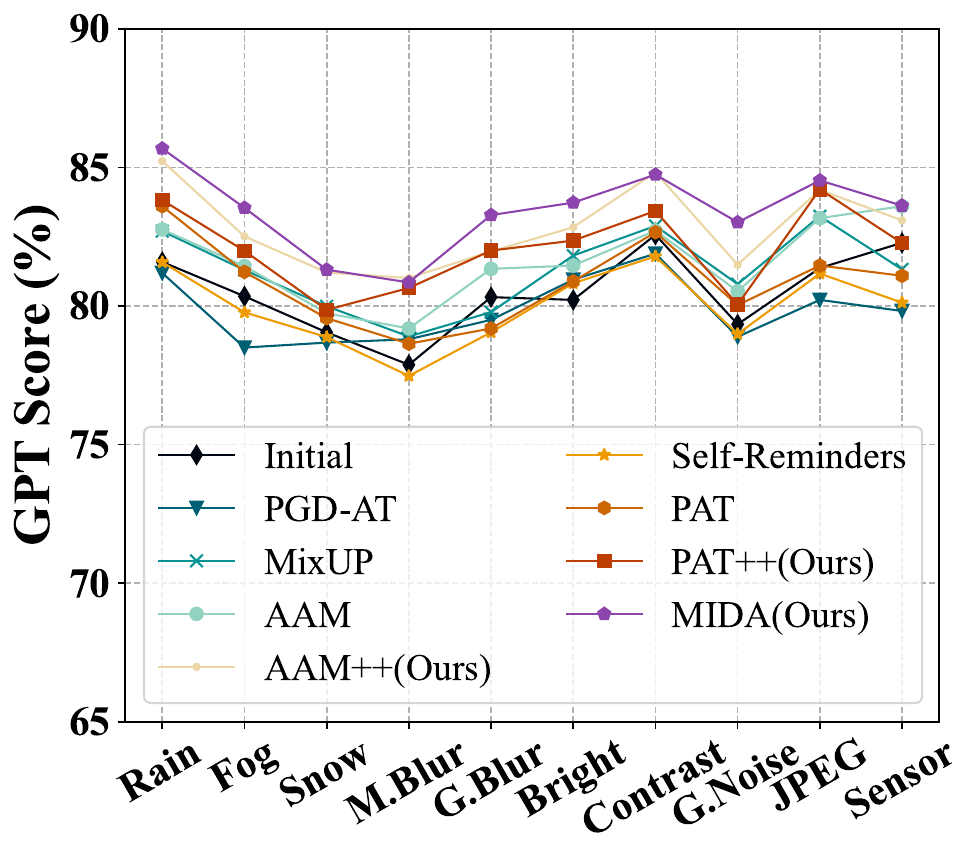}%
        \label{fig:drivelm_nonillusion}%
    }

    \caption{\revised{Evaluation of noise defense methods. For the first 4 subfigures, the x-axis denotes different types of illusions. The last 2 subfigures ($^\dagger$) report the performance of LaneATT and DriveLM under different non-illusion perturbations.}}
    \label{fig:enhance}
    \vspace{-0.1in}
\end{figure}

\textbf{Enhancement for conventional LD models.} In this case, since MIDA only uses the image modality, it effectively corresponds to AAM++. Specifically, we adopt ResNet18 for LaneATT and UFLD, and apply a range of robustness enhancing techniques including PGD adversarial training \cite{madry2017towards_pgd}, cutout \cite{devries2017improved_cutout}, copy paste \cite{ghiasi2021simple_copypaste}, HSV augmentation, and MixUP \cite{zhang2017mixup}, all known to improve robustness against adversarial noise \cite{liu2019perceptual, liu2020bias, liu2023towards, liu2023x} and natural corruptions. We also include AAM \cite{zhang2024lanevil} as a baseline. \Fref{fig:laneatt} and \Fref{fig:ufld} show that AAM++ achieves the highest accuracy gain of 4.23\% on average, building upon the already strong performance of AAM. In contrast, other methods provide only modest improvements below 1.5\%, with an average gain of 1.24\%.

\textbf{Enhancement for \advlmns.} We use DriveLM and Dolphins as base models and apply a series of robustness enhancing techniques, including visual modality methods such as PGD adversarial training \cite{madry2017towards_pgd}, MixUP \cite{zhang2017mixup}, AAM \cite{zhang2024lanevil}, and AAM++; textual modality methods such as PAT \cite{mo2024fight}, Self-Reminders \cite{xie2023defending}, and PAT++; as well as the cross modality method MIDA. As shown in \Fref{fig:drivelm} and \Fref{fig:dolphins}, our MIDA and its components (AAM++ and PAT++) achieve the highest GPT score gain of 3.82\%, outperforming other visual, textual, and cross-modality methods, which average a gain of 1.93\%.

\revised{\textbf{Robustness under non-illusion perturbations.}
We further evaluate MIDA under non-illusion corruptions such as weather, blur, brightness/contrast shifts, noise, JPEG compression, and sensor artifacts. As shown in \Fref{fig:laneatt_nonillusion} and \Fref{fig:drivelm_nonillusion}, MIDA consistently achieves the highest performance across all perturbations. For LaneATT, the Accuracy remains within the 74–88\% range and exceeds the other methods by 1–2\% on several perturbations (\eg, Snow, Gaussian Blur). For DriveLM, GPT Score remain within 78–86\%, with MIDA achieving the highest results and exhibiting only small fluctuations (typically below 1\%). These results confirm that MIDA enhances generalizable robustness rather than inducing illusion-specific overfitting.}

\subsection{Visual Fidelity Analysis}
\label{sec:visual-fid}
In this part, we further study the visual fidelity of our generated images. Specifically, we conduct three experiments and analysis: (1) training on either simulated or real-world datasets and then testing on another dataset; (2) conducting human perception and (3) conducting ADVLM perception studies on the visual quality of our \toolns.


\textbf{Cross-domain LD model prediction.}
\ding{182} \emph{LanEvil $\rightarrow$ Real-world.} For each of the three real-world datasets (TuSimple, CULane, LLAMAS), we separately train a model on the original real-world dataset, a model on the generated \toolns, and a \tool pre-trained model fine-tuning on 100 images from the corresponding real-world datasets. All the models are LaneATT with ResNet-34. Here, we report the F1-score (\%) on the testing set of TuSimple (96.77, 91.56, 94.23), CULane (76.68, 69.32, 73.98), and LLAMAS (93.74, 89.15, 92.58). The results indicate acceptable domain gaps despite differences between the real world and our simulated images. Additionally, the gap can be narrowed by incorporating a small number of real images. \ding{183} \emph{Real-world $\rightarrow$ LanEvil.} Furthermore, we use LaneATT models with ResNet-34 and train them on real-world datasets (\ie, TuSimple, CULane, and LLAMAS), and then w/ or w/o fine-tuning them on the \toolns. For each model, we test it separately on corresponding real-world datasets and \toolns. \Tref{tab:fine_tuning_results} shows minor performance reductions, indicating a comparatively small domain gap between \tool and other real-world datasets.

\revised{
Although the cross-domain results show only a minor 2–5\% drop in F1-score, these differences reveal that simulation-generated illusions may still deviate from real-world patterns in texture, scale, or contextual lighting. 
Such gaps could introduce a mild bias in visual statistics, limiting the model’s ability to generalize to unseen real-world environments. Nevertheless, simulation-based evaluation remains a widely adopted and cost-efficient paradigm in AD testing, as it enables controlled, repeatable, and low-risk experimentation. With continuous advances in physically based rendering, neural material synthesis, and domain randomization, this gap is expected to further narrow, enhancing both the realism and reliability of future evaluations.
}


\begin{table}[]
\centering
\renewcommand\arraystretch{0.95}
\small

\caption{F1-Scores (\%) of LaneATT with ResNet-34 w/o and w/ fine-tuning on \tool training set.}

\label{tab:fine_tuning_results}
\resizebox{1.0\linewidth}{!}{
\begin{tabular}{@{}c|cc>{\columncolor[HTML]{EFEFEF}}c|cc>{\columncolor[HTML]{EFEFEF}}c@{}}
\toprule
\multicolumn{1}{c|}{}                                   & \multicolumn{3}{c|}{Real-world test set}                                        & \multicolumn{3}{c}{\tool test set} \\
\cmidrule(l){2-7} 
\multicolumn{1}{c|}{\multirow{-2}{*}{\textbf{Dataset}}} & w/o & w/ & {\color[HTML]{333333} Gap} & w/o  & w/  & Gap \\ \midrule
TuSimple \cite{tusimple}                                                & 96.77  & 95.40  & -1.37                          & 48.34   & 62.92  & +14.58    \\
CULane \cite{pan2018spatial_SCNN}                                                  & 76.68  & 76.14  & -0.54                          & 55.98   & 68.41  & +12.43    \\
LLAMAS \cite{behrendt2019unsupervised_llamas}                                                & 93.74  & 93.09  & -0.65                          & 58.26   & 66.95  & +8.69     \\ \bottomrule
\end{tabular}
}
\vspace{-0.1in}
\end{table}

\textbf{Human perception study.} Following \cite{Li_2023_CVPR}, we conducted a human perception study with 100 participants from campus, all with normal (corrected) vision. They evaluated the naturalness of 300 images (150 from \tool and 150 from real-world scenarios). Each participant viewed an image for 3 seconds and then rated it using a 5-point Absolute Category Rating (ACR) \cite{hosu2020koniq_ACR}. The entire evaluation was completed in 30 minutes. The average ACR results (our images: 3.89 and real-world images: 3.98) suggest that our simulated images are comparatively natural to human vision when compared to real-world images. 

\textbf{ADVLM perception study.} Following the human perception study, we further evaluate visual fidelity from the perspective of \advlmns. We use both Dolphins and DriveLM as evaluation models on 300 images, including 150 from \tool and 150 from real-world scenarios. The evaluation metric is the ASR. The average ASR is 4.23 for \tool and 4.32 for real-world images, indicating that the visual quality of \tool is comparable to real-world inputs from the perspective of ADVLM perception.

\subsection{\revised{Combined Illusions Analysis}}
\label{sec:combined-illusions}

\revised{
To further investigate the combined influence of multiple environmental illusions, we conduct a pilot study.
Based on the four illusion categories, namely Road Damage (RD), Traffic Obstruction (TO), Shadow (SH), and Reflection (RF), we systematically construct six different two-illusion combinations.
Each combination contains 100 images for conventional LD evaluation and 20 clips for \advlm testing, created by sequentially applying both illusions within the same simulated environment.
As this experiment emphasizes analysis under single controlled factors rather than comprehensive benchmark expansion, the results provide a concise quantitative reference for assessing cross-illusion interaction and compositional effects.
}

\begin{table}[t]
\centering
\caption{\revised{Performance under combined illusions. ``Gap'' denotes the performance drop compared to the original non-illusion condition. Results are reported for GANet under Accuracy (\%) and DriveLM under GPT Score (\%).}}
\vspace{2mm}

\begin{tabular}{@{}c@{}}
\label{tab:combine_GANet}
\subfloat[Results for GANet under Accuracy (\%)]{
\huge
\resizebox{0.49\textwidth}{!}{%
\begin{tabular}{c|cccccc}
\toprule[1.4pt]
\textbf{Metric} & \textbf{RD+TO} & \textbf{RD+SH} & \textbf{RD+RF} & \textbf{TO+SH} & \textbf{TO+RF} & \textbf{SH+RF} \\ 
\midrule
Original  & 81.5 & 82.1 & 82.4 & 81.0 & 81.3 & 82.6 \\
Perturbed & 72.9 & 73.0 & 74.1 & 71.6 & 72.5 & 73.4 \\
\rowcolor[HTML]{EFEFEF}
Gap       & -8.6 & -9.1 & -8.3 & -9.4 & -8.8 & -9.2 \\
\bottomrule[1.4pt]
\end{tabular}
}}\end{tabular}

\begin{tabular}{@{}c@{}}
\label{tab:combine_OmniL}
\subfloat[Results for DriveLM under GPT Score (\%)]{
\huge
\resizebox{0.49\textwidth}{!}{%
\begin{tabular}{c|cccccc}
\toprule[1.4pt]
\textbf{Metric} & \textbf{RD+TO} & \textbf{RD+SH} & \textbf{RD+RF} & \textbf{TO+SH} & \textbf{TO+RF} & \textbf{SH+RF} \\
\midrule
Original  & 79.5 & 79.8 & 80.1 & 78.9 & 79.2 & 80.3 \\
Perturbed & 76.0 & 75.6 & 76.7 & 73.9 & 74.8 & 76.2 \\
\rowcolor[HTML]{EFEFEF}
Gap       & -3.5 & -4.2 & -3.4 & -5.0 & -4.4 & -4.1 \\
\bottomrule[1.4pt]
\end{tabular}
}}\end{tabular}

\label{tab:combined}
\vspace{-0.1in}
\end{table}

\revised{
As shown in \Tref{tab:combined}, the results from this small-scale experiment suggest that combining two types of environmental illusions leads to a stronger performance degradation compared with individual factors, although the increase remains moderate.
For the conventional model (GANet), the average Accuracy decreases from 7.5\% under single illusion conditions to 8.9\% when two illusions appear simultaneously.
For the \advlm (DriveLM), the GPT Score decreases from 2.3\% to 4.1\%.
These findings imply that concurrent environmental illusions indeed aggravate recognition difficulty, but the overall influence grows within a limited range rather than explosively. \textit{Experiment of overlaying more environmental illusions is in the Supplementary Materials.}}

\subsection{\revised{Parameter Sensitivity Analysis}}
\label{sec:parameter_sensitivity}

\revised{We further analyze the influence of MIDA’s main hyperparameters using one representative conventional LD model (LaneATT for AAM++) and one representative ADVLM (DriveLM for PAT++).  
Each parameter was varied over a wide range.  
\Tref{tab:ablation} indicates that all metrics exhibit only minor fluctuations across different settings.  
The final configurations ($T_{HAA}{=}0.6$, $T_{LAA}{=}0.4$, $\beta{=}0.5$, $\alpha{=}0.6$) are chosen as they provide consistently stable performance plateaus for both conventional LD models and ADVLMs, avoiding overfitting to specific hyperparameter values.}

\begin{table}
\centering
\caption{\revised{Parameter Sensitivity Experiments Results}}
\label{tab:ablation}

\begin{tabular}{@{}c@{}}
\label{tab:sens_thaa}
\subfloat[AAM++ sensitivity to $T_{HAA}$ on LaneATT.]{
\resizebox{0.49\textwidth}{!}{%
\begin{tabular}{c|cccccc}
\toprule
\textbf{Metric} & 0.2 & 0.3 & 0.4 & 0.5 & 0.6 & 0.7 \\
\midrule
Accuracy (\%)  & 89.12 & 89.25 & 89.41 & 89.58 & \textbf{89.73} & 89.57 \\
F1-Score (\%)  & 83.42 & 83.55 & 83.71 & 83.89 & \textbf{84.02} & 83.87 \\
\bottomrule
\end{tabular}
}}\end{tabular}

\begin{tabular}{@{}c@{}}
\label{tab:sens_tlaa}
\subfloat[AAM++ sensitivity to $T_{LAA}$ on LaneATT.]{
\resizebox{0.49\textwidth}{!}{%
\begin{tabular}{c|cccccc}
\toprule
\textbf{Metric} & 0.05 & 0.1 & 0.2 & 0.3 & 0.4 & 0.5 \\
\midrule
Accuracy (\%)  & 88.98 & 89.14 & 89.32 & 89.47 & \textbf{89.62} & 89.45 \\
F1-Score (\%)  & 83.28 & 83.44 & 83.59 & 83.75 & \textbf{83.93} & 83.78 \\
\bottomrule
\end{tabular}
}}\end{tabular}

\begin{tabular}{@{}c@{}}
\label{tab:sens_beta}
\subfloat[AAM++ sensitivity to mixing ratio $\beta$ on LaneATT.]{
\resizebox{0.49\textwidth}{!}{%
\begin{tabular}{c|cccccc}
\toprule
\textbf{Metric} & 0.1 & 0.2 & 0.3 & 0.4 & 0.5 & 0.6 \\
\midrule
Accuracy (\%)  & 89.21 & 89.33 & 89.49 & 89.61 & \textbf{89.78} & 89.74 \\
F1-Score (\%)  & 83.48 & 83.60 & 83.72 & 83.83 & \textbf{84.01} & 83.96 \\
\bottomrule
\end{tabular}
}}\end{tabular}

\begin{tabular}{@{}c@{}}
\label{tab:sens_alpha}
\subfloat[PAT++ sensitivity to balancing coefficient $\alpha$ on DriveLM.]{
\resizebox{0.49\textwidth}{!}{%
\begin{tabular}{c|cccccc}
\toprule
\textbf{Metric} & 0.1 & 0.2 & 0.3 & 0.4 & 0.6 & 0.8 \\
\midrule
GPT Score (\%)      & 80.49 & 80.61 & 80.69 & 80.78 & \textbf{80.93} & 80.71 \\
Language Score (\%) & 30.09 & 30.22 & 30.33 & 30.41 & \textbf{30.54} & 30.46 \\
\bottomrule
\end{tabular}
}}\end{tabular}
\vspace{-0.1in}
\end{table}

\revised{All metrics exhibit only minor variations (within approximately 0.6\% for AAM++ and 0.3\% for PAT++), which confirms the overall stability of the MIDA.  
The final parameter configuration is selected at the point where both accuracy and robustness reach stable and consistent plateaus.}

\section{Simulation Case Studies}\label{sec:case_study}
We conduct closed-loop evaluations in a simulation environment (\ie, CARLA) using two models: OpenPilot \cite{openPilot} for conventional LD models and LMDrive \cite{shao2024lmdrive} for \advlmns. The 3D scenarios from \tool are used as direct perception inputs, and the systems are evaluated based on their final decision-making performance.

\subsection{OpenPilot Simulation}
\label{sec:5.1}
We begin by evaluating \tool on OpenPilot, an open-source commercial driver assistance system that has been deployed in real-world vehicles such as those from TOYOTA. Eight cases are selected from \toolns, covering two types under each illusion category. Each case is configured with a defined starting point and direction to ensure traversal through the designated road sections. The evaluation pipeline is as follows: \ding{182} connect OpenPilot with a source-compiled CARLA 0.9.14 simulator and configure initialization parameters such as maps; \ding{183} for each clean and perturbed 3D case pair, choose a vehicle model (\ie, Audi) and set the starting position; \ding{184} enable autonomous driving mode with a speed limit of 45 mph and assess the system’s decision-making performance. The starting point is set approximately 5–10 \emph{meters} before the illusion region.

\begin{figure}[!t]
    \centering
    \subfloat[Clean Scenario]{\includegraphics[width=0.96\linewidth]{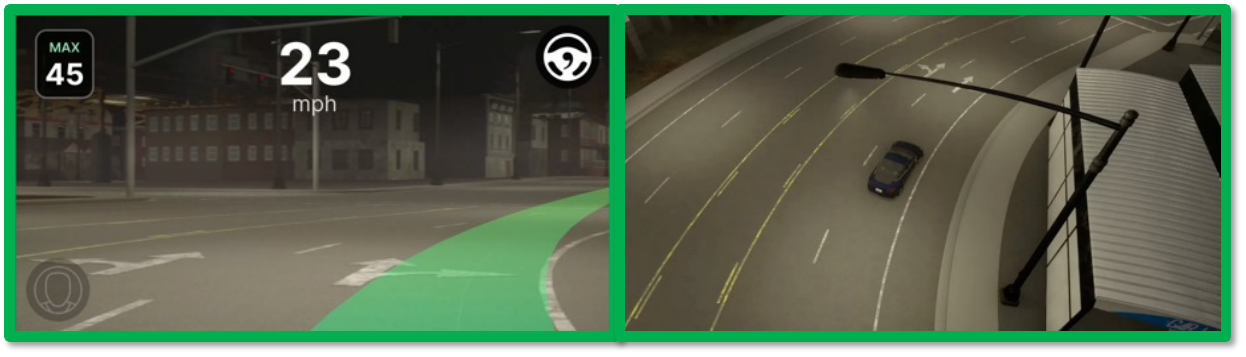}
        \label{subfig:openPilot1}
    }
   
    \subfloat[Perturbed Scenario]{
        \includegraphics[width=0.96\linewidth]{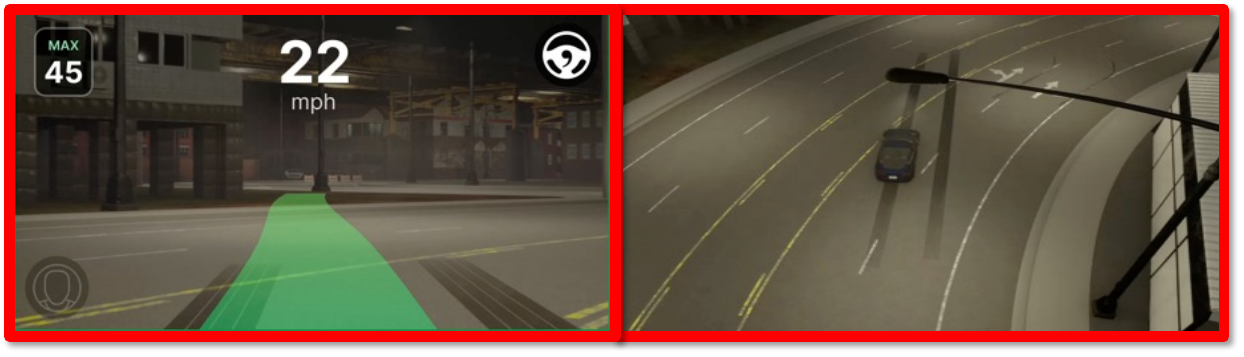}
        \label{subfig:openPilot2}
    }
    \caption{The \texttt{Tire Marks} case causes OpenPilot to make incorrect decisions leading to collisions on the wall.}
    \label{fig:openPilot1}
\end{figure}


We repeat each experiment five times across all eight cases and report the average results. Following \cite{sato2021dirty}, we use Attack Success Rate (ASR) as the evaluation metric. According to the US traffic policy \cite{aashto2001policy_US}, a perturbation is considered successful if it causes the vehicle to deviate laterally by more than 0.285 meters within 2.5 seconds. As driving time progresses, we observe a significant increase in ASR across all types of illusions, indicating that the vehicle increasingly makes incorrect decisions. For example, in 92.31\% of frames, the \texttt{Road Damage} scenario causes OpenPilot to deviate by more than 0.285 meters within the required timeframe. As shown in \Fref{fig:openPilot1}, the \texttt{Tire Marks} scenario leads to frequent recognition failures, ultimately causing the vehicle to collide with the wall. These findings demonstrate that the proposed environmental illusions substantially compromise the robustness of commercial AD systems.

\subsection{LMDrive Simulation}
\label{sec:5.2}
After evaluating OpenPilot, which is based on conventional LD models, we further assess the performance of \advlm using LMDrive \cite{shao2024lmdrive}, a representative ADVLM, in the CARLA simulator. The evaluation procedure is similar to that of OpenPilot: \ding{182} start the Docker-based CARLA 0.9.10.1 environment; \ding{183} launch the CARLA leaderboard with the selected agent (\ie, Audi); and \ding{184} enable driving mode to begin evaluation. The starting point is set 5–10 \emph{meters} before the illusion region.

\begin{figure}
    \centering
    \includegraphics[width=0.97\linewidth]{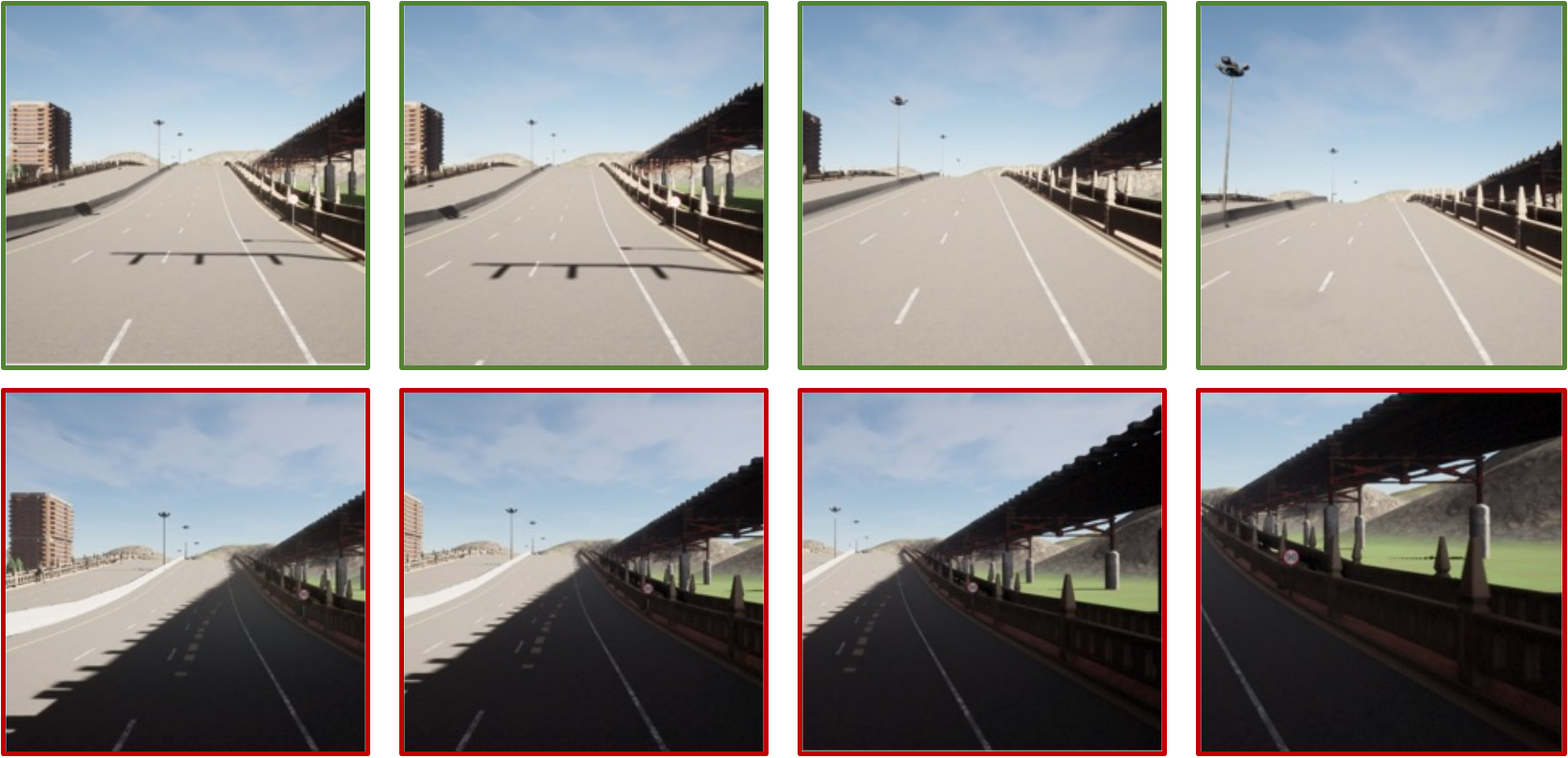}
    \caption{The \texttt{Rail} case causes LMDrive to make incorrect decisions leading to collision on the road side.}
    \label{fig:LMDrive}
    \vspace{-0.1in}
\end{figure}

We select eight cases for evaluation, with each experiment repeated five times. The evaluation metric is ASR, following the details outlined in \Sref{sec:5.1}. Results show that ASR reaches 77.5\% in environments with illusions, compared to only 11.25\% in environments without illusions. As shown in \Fref{fig:LMDrive}, the \texttt{Rail} environmental illusion leads the vehicle controlled by LMDrive to collide.

In summary, the above evaluations on both systems highlight the potential threats posed by the proposed environmental illusions to closed-loop AD systems, whether based on conventional LD models or \advlmns. These findings raise safety concerns for real-world AD vehicles.

\section{Real-world Case Studies}
\label{sec:realworld}
Finally, we extend our closed-loop experiments from simulation to real-world scenarios to evaluate the impact of environmental illusions in practical settings.
We use two AD test vehicles: \ding{182} Jetbot \cite{jetbot}, an open-source platform with an NVIDIA Jetson Nano and an RGB camera, features two-wheel differential drive for basic motion control. \ding{183} LIMO PRO \cite{LIMO}, equipped with an NVIDIA Orin Nano, LiDAR, and a depth camera, adopts a four-wheel chassis with Ackermann steering for realistic vehicle dynamics. Due to the characteristics of both vehicles, we tested the conventional LD models on Jetbot vehicles and \advlm on LIMO vehicles.

The experiment was conducted on an AGILE•X sand table. \revised{Four representative routes were selected from \tool and reconstructed on the sand table, covering straight segments, turning curves, and mixed conditions. Two illusion variants were applied to every route: Shadow scenes generated by adjustable overhead lighting and Road Damage scenes emulated with textured surface patches. This setup ensured tight spatial and visual alignment with the simulated scenarios.} Each scenario was tested 10 times. A trial is considered successful if the vehicle completes the route without a collision. The evaluation metric is calculated as the ratio of successful trials to the total number of trials.

\begin{table}[t]
\centering
\caption{\revised{Performance comparison between simulated and real-world routes. Each real route replicates the corresponding simulated layout in \toolns. The Conventional LD Model is evaluated by Accuracy (\%), and the \advlm by the GPT Score (\%).}}
\resizebox{\linewidth}{!}{
\begin{tabular}{@{}c|cc|cc@{}}
\toprule[0.8pt]
\multirow{2}{*}{\textbf{Route Type}} & \multicolumn{2}{c|}{LaneATT} & \multicolumn{2}{c}{Dolphins} \\ \cmidrule(l){2-5} 
 & Simulated & Real-world & Simulated & Real-world \\ \midrule
Straight & 72.4 & 70.9 & 74.1 & 72.8 \\
Right Turn & 71.6 & 70.3 & 73.8 & 72.1 \\
Left Turn & 69.8 & 68.4 & 71.2 & 70.1 \\
Mixed Conditions & 70.5 & 68.9 & 72.0 & 70.7 \\ \midrule
Average & 71.1 & 69.6 & 72.8 & 71.4 \\ \bottomrule[0.8pt]
\end{tabular}
}
\label{tab:real_route_eval}
\end{table}

\revised{\textbf{Sim2Real Consistency Validation.} To verify visual consistency, 50 frames were sampled at matched positions from both domains for LD. We chose representative models (\ie, LaneATT for conventional LD models and Dolphins for \advlmns). As shown in \Tref{tab:real_route_eval}, the LaneATT achieved an average Accuracy of 71.1\% in simulation and 69.6\% in real tests, while the Dolphins reached GPT Score of 72.8\% and 71.4\%. The close results indicate that both models exhibit stable performance when transferred from simulation to real environments, confirming the fidelity of scene replication and the validity of the evaluation.}

\begin{table}[!t]
\caption{The experimental results of real-world case studies. The values are shown in \emph{success times / total times}.}
\label{tab:real-exp}
\resizebox{\linewidth}{!}{
\centering
\begin{tabular}{@{}c|cccc|cccc@{}}
\toprule[1.0pt]
\multirow{2}{*}{\textbf{}} & \multicolumn{4}{c|}{\textbf{JetBot} \cite{jetbot}} & \multicolumn{4}{c}{\textbf{LIMO} \cite{LIMO}} \\ \cmidrule(l){2-9} 
 & Route 1 & Route 2 & Route 3 & Route 4 & Route 1 & Route 2 & Route 3 & Route 4 \\ \midrule
w/o illusions & 8 / 10 & 7 / 10 & 8 / 10 & 9 / 10 & 6 / 10 & 5 / 10 & 5 / 10 & 7 / 10 \\ \midrule
w/ illusions & 2 / 10 & 2 / 10 & 4 / 10 & 1 / 10 & 0 / 10 & 1 / 10 & 2 / 10 & 1 / 10 \\ \bottomrule[1.0pt]
\end{tabular}
}
\end{table}

\subsection{Jetbot Vehicle Evaluation}
We deployed the LaneATT model with a ResNet34 backbone on the Jetbot vehicle. The model's output was used to detect lane markings and compute the lane center. Steering decisions were made based on the center's horizontal position in the image: left area (0–40\%) triggers a right turn, center area (40–60\%) drives straight, and right area (60–100\%) triggers a left turn. The control loop runs every 0.5 seconds, and the vehicle speed is set to 0.2 m/s. As shown in the left part of \Tref{tab:real-exp}, in the scenario without environmental illusions, the JetBot achieves a success completion rate (SCR) of 80.00\%. However, in the presence of environmental illusions, the SCR drops significantly to 22.5\%. This indicates that environmental illusions continue to pose a significant threat on conventional LD models in real-world scenarios. The \Fref{subfig:rr1} and \Fref{subfig:rr2} show the performance of the Jetbot vehicle under \texttt{Shadow} illusions. \revised{We further inspect representative failure cases on JetBot to better understand the failure modes caused by environmental illusions. Under Shadow and Reflection illusions, critical lane boundaries become partially occluded, causing LaneATT to misidentify the lane center and issue incorrect turning commands.}

\subsection{LIMO Vehicle Evaluation}
\begin{figure}[!t]
    \centering
    \subfloat[Clean Scenario for Jetbot Evaluation]{\includegraphics[width=0.96\linewidth]{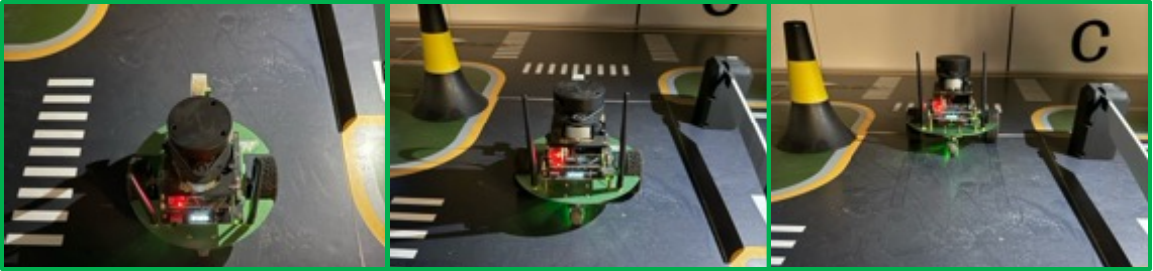}
        \label{subfig:rr1}
    }
   
    \subfloat[Perturbed Scenario for Jetbot Evaluation]{
        \includegraphics[width=0.96\linewidth]{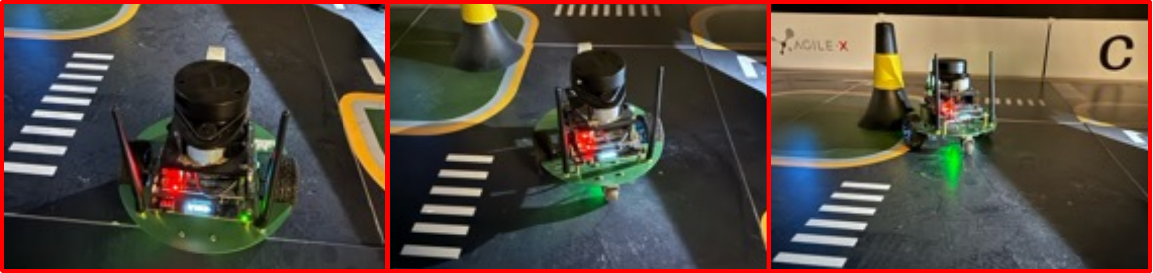}
        \label{subfig:rr2}
    }

    \subfloat[Clean Scenario for LIMO Evaluation]{\includegraphics[width=0.96\linewidth]{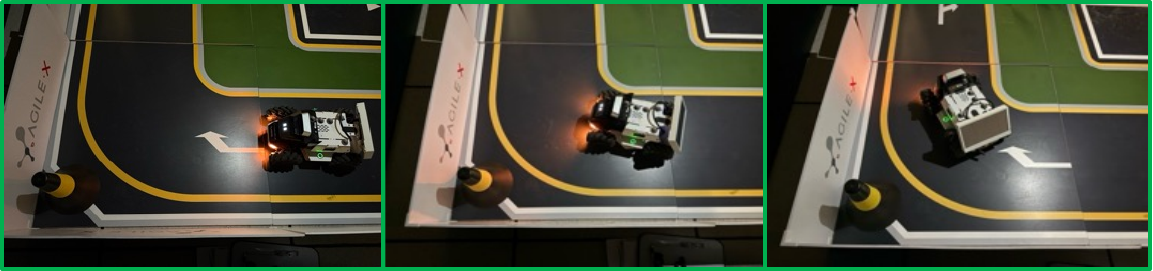}
        \label{subfig:rr3}
    }
   
    \subfloat[Perturbed Scenario for LIMO Evaluation]{
        \includegraphics[width=0.96\linewidth]{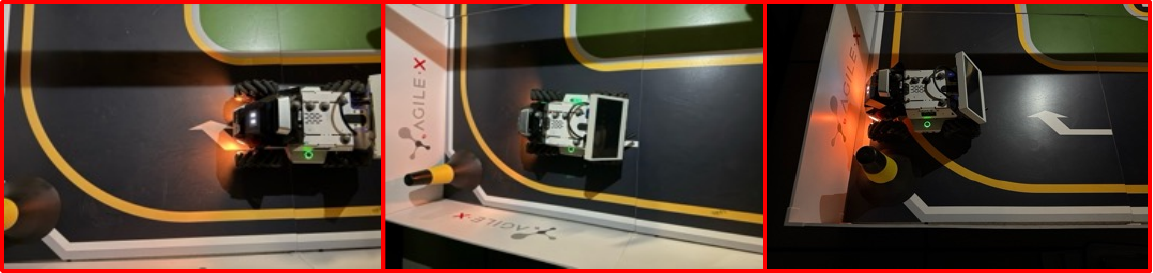}
        \label{subfig:rr4}
    }
    
    \caption{Consecutive images showing the LD model–driven Jetbot and the ADVLM–driven LIMO failing to interpret the required actions when confronted with \texttt{Shadow} environmental illusions, ultimately resulting in collisions.}
    \label{fig:real-world-results}
\end{figure}

We deploy \advlm on LIMO vehicles by integrating two large models, Dolphins and LLaMA, as described in \cite{zhang2024visual,wang2025black}. This setup enables the translation of high-level driving commands into low-level control signals suitable for real-world execution. As shown in the right part of \Tref{tab:real-exp}, the system achieves an SCR of 57.5\% without environmental illusions. However, when environmental illusions are introduced, the SCR drops significantly to 10.0\%, further highlighting the strong impact of such illusions on \advlmns. The \Fref{subfig:rr3} and \Fref{subfig:rr4} show the performance of LIMO under the \texttt{Shadow} environmental illusion. \revised{Qualitative inspection of LIMO failure cases reveals a different failure pattern from JetBot. When Shadow illusions obscure high-contrast cues, the visual encoder provides weakened features, leading DriveLM to generate semantically plausible but operationally incorrect commands.}

\section{Conclusions and Future Work}

This paper investigates the potential threats posed by environmental illusions in autonomous driving (AD) from a perspective of lane perception. We establish \toolns, the first comprehensive benchmark for evaluating model robustness under such conditions. Large-scale experiments show that naturally occurring environmental illusions significantly degrade the performance of lane perception models, underscoring the need for greater attention to this overlooked challenge in building reliable AD systems. The proposed MIDA framework further improves robustness against these illusions. We hope this study raises awareness of the threat of environmental illusions and encourages further research toward more robust and reliable AD system. To access the dataset, please contact the authors via email.

\textbf{Limitations.} Despite the promising results, several areas remain to be explored:  \ding{182} evaluation of \tool on other common tasks like 3D detection; and \ding{183} testing more product-level real-world autonomous driving vehicles.
\section{Acknowledgments}
This work was supported by the National Natural Science Foundation of China (Grant. 62476018),  Beijing Natural Science Foundation (Grant. QY24136).

\bibliographystyle{IEEEtran}
\bibliography{sample-base}

\end{document}